%% file: jansen22structuredpreprint.tex
\documentclass[11pt,a4paper]{article}
  
\usepackage{latexsym}
\usepackage{graphicx}
\usepackage{helvet}
\usepackage{psfrag}
\usepackage{bbm}
\usepackage{bm}
\usepackage{setspace}
\usepackage{comment}

\usepackage{natbib}
\usepackage{amsmath}  
\usepackage{mathabx}  
\usepackage{multirow}
\usepackage{url}

\renewcommand{\vec}{\bm}

\newcommand{\norm}[1]{\| #1 \|}
\newcommand{\floor}[1]{\left\lfloor #1\right\rfloor}

\newcommand{\RR}{\mathbbm{R}}

\newcommand{\sign}{\mathrm{sign}}
\newcommand{\var}{\mathrm{var}}

\renewcommand{\emptyset}{\mbox{\O}}
\def\Chi{%
    \mbox{
    {\kern-.20em\setbox0=\hbox{X}%
    \vbox to 1.0\ht0{\hbox{$\chi$}\vss}}%
    \kern-.00em} }
\def\mathlatex{%
    \mbox{
    L\kern-.36em {\setbox0=\hbox{T}%
    \vbox to \ht0{\hbox{\the\scriptfont0 A}\vss}}%
    \kern-.15em \TeX} }

\newcommand{\qed}{\hbox{}\hfill $\Box$ }

\newtheorem{assump}{Assumption}[section]

\newenvironment{blinded}{}{}

\newenvironment{keywords}{\noindent\textbf{Keywords:}}{}
\renewcommand{\and}{;}

\excludecomment{twocolversion}

\newcommand{\papertitle}{Information criteria for structured parameter
selection in high dimensional tree and graph models}

\title{\papertitle}
\date{\today}
\begin{blinded}
\author{Maarten Jansen\\
Universit\'e libre de Bruxelles, departments of Compter Science and Mathematics}
\end{blinded}

\begin{document}

\maketitle

\input jansen22structuredabstract.tex
\input jansen22structuredkeywords.tex
\newpage
\input jansen22structuredbody1or2col.tex
\bibliographystyle{plainnat}
\bibliography{jansen22structuredpreprint}

\end{document}

%% file: jansen22structuredabstract.tex
\begin{abstract}
Parameter selection in high-dimensional models is typically finetuned in a way
that keeps the (relative) number of false positives under control. This is
because otherwise the few true positives may be dominated by the many possible
false positives. This happens, for instance, when the selection follows from
a naive optimisation of an information criterion, such as AIC or Mallows's Cp.
It can be argued that the overestimation of the selection comes from the
optimisation process itself changing the statistics of the selected variables,
in a way that the information criterion no longer reflects the true divergence
between the selection and the data generating process. In lasso, the
overestimation can also be linked to the shrinkage estimator, which makes the
selection too tolerant of false positive selections. For these reasons, this
paper works on refined information criteria, carefully balancing false
positives and false negatives, for use with estimators without shrinkage. In
particular, the paper develops corrected Mallows's Cp criteria for structured
selection in trees and graphical models. 
\end{abstract}

%% file: jansen22structuredkeywords.tex

%% file: jansen22structuredbody1or2col.tex
\section{Introduction}
\label{sec:intro}

Whereas statistical inference takes place after the estimation of the
parameters in a model, model and parameter selection is a step that in
principle precedes  and in some cases includes the estimation of the
parameters.
Statistical inference operates in an asymmetric way w.r.t.~the null and
alternative hypotheses, thereby reflecting the presumption of innocence.
In parameter selection, the decisions whether or not to include parameters
in the model are based on symmetric criteria, which can be compared to
profiling in a criminal investigation.

In all three steps, parameter selection, estimation and inference, the notion of
likelihood may be adopted, although each time in a different role.
In inference, the likelihood ratio can be used to test the significance of a
larger model against a smaller.
An estimation procedure assumes that the model being worked in is correct, or
at least it has been selected as the least false model to work in for the
statistical inference problem at hand.
In this framework, maximising the likelihood function amounts to finding the
parameters that make the data fit as well as possible into the proposed model.
In a parameter selection, candidate models can be assessed through the
capability of modelling or predicting the same sort of data as those that have
been observed. Working with maximal likelihood would promote large models, 
because they would best fit the current observations, including the noise. 
The best model for predicting these observations, without noise that is, would
minimise the \emph{expected} value of the maximum likelihood with respect to
the data generating process (DGP, also referred to as ground truth).
The expected likelihood is closely related to the Kullback-Leibler (KL)
divergence between the proposed model and the DGP.
Other, similar divergences between the proposed model and the DGP include the
prediction error (PE). PE is basically the KL divergence in a model with
additive normal noise, assuming the variance to be known (or easy to estimate
outside the parameter selection).

The divergence (i.e., the expected likelihood) depends on the bias and the
variance of an estimator in the proposed model.
The balance between bias and variance is typically equivalent,
at least in expected value, to a closeness-complexity trade-off, formalised in
an information criterion.
Closeness is typically expressed by the sample likelihood in the model under
consideration, while the complexity acts as a penalty
compensating for the gap between sample likelihood and the expected likelihood.
Akaike's Information Criterion (AIC)
\citep{akaike73:AIC} and Mallows's $C_p$ \citep{mallows73:Cp} are prototypes of
these information criteria, estimating, respectively, the KL distance and the
PE.

In the last few decades, the interaction between parameter selection and
statistical inference has become a major point of attention in statistical 
research.
In one direction, the focus for which the model is used determines which model
performs best \citep{claeskens03:FIC}.
In the other direction, the parameter selection procedure creates additional
uncertainty in the estimators, leading to wider confidence intervals
\citep{berk13:validpostselectioninference,vandegeer14:desparsifiedlasso,zhang14:desparsifiedlasso,lee16:exactpostselectioninference,charki18:AICpostselection}.

This paper concentrates on the uncertainty arising from the selection of
variables, however not on its effect on the subsequent inference. Instead, it
deals with its effect on the statistics of the information criterion used in
the process of parameter selection. Indeed, the classical information criteria
have been developed for estimating the divergence of a given, fixed model from
the DGP. However, if the model under consideration comes from optimising a
criterion, that optimisation has interacted with the noise. This interaction
affects the statistics of the noise.
As a result, the information criterion may not be a good estimator
of the divergence after all, and hence, the optimiser of the criterion may
point to a suboptimal model in terms of divergence w.r.t.~the DGP.
The gap between the information criterion and the divergence or error measure
can be explained and formalised by the concept of generalised degrees of
freedom \citep{ye98:modelselection,hansen14:dof,jansen14:ICsparsity}.

This paper studies the generalised degrees of freedom in the case of
structured parameter selection in trees and graphs. 
It starts off with a lasso procedure, i.e., a selection and estimation by
solving an $\ell_1$ regularised least squares problem
\citep{tibshirani96:lasso,chen98:basispursuit}. 
It has been reported
\citep{wainwright09:sharpthresholds,tropp06:justrelax,zhao06:lassoconsistency}
that lasso is selection consistent, at least if the DGP
is described by a model belonging to the set of models under
consideration. Moreover, the parameters in that model are assumed to be
sufficiently large, while at the same time the regularisation parameter
$\lambda$ should not be too large, so that for $n \to \infty$ the numbers of
false positives and false negatives tend to zero.
As these assumptions put quite some restrictions on $\lambda$, it is generally
impossible to combine minimum prediction errors and selection consistency
\citep{meinshausen06:graphs}, which motivates the use of adaptive lasso
\citep{zou06:adaptivelasso}. In particular, a minimum prediction error choice
of $\lambda$ leaves many false positives resulting in noisy features. This
effect can be explained by the lasso shrinkage in two ways.
First, the shrinkage reduces the price of false positives in terms of induced
variance. As a result, the optimisation of the bias-variance balance is
tolerant of their presence in the selection.
Second, the shrinkage is responsible for an important bias in the lasso
estimator. In order to keep that shrinkage bias under control, the minimum
prediction error tends to be achieved for small values of the regularisation
$\lambda$, which corresponds to large models.
Debiasing \citep{javanmard18:debiasinglasso} or regularisation
\citep{li15:rlasso} of the lasso solution leads to minimum prediction error
selections with less false positives.
Alternative regularisations have also been proposed, reducing the shrinkage in
large significant parameters.
They include the smoothly clipped absolute deviation (SCAD) \citep{fan01:scad}
and a minimax concave penalty (MCP) \citep{zhang10:mcp}.

This paper follows a debiasing approach. It adopts the lasso algorithm,
minimising an $\ell_1$ regularised sum of squared residuals, for selection
purposes only, not for estimation. The built in biased, shrinkage estimation is
replaced by a least squares projection onto the selected model.
The use of lasso for selection purposes is motivated by the fact that for a
given value of $\lambda$, the convex optimisation of lasso leads to
nearly the same sparsity level as a combinatorially complex $\ell_0$
regularisation, leading to the best $\kappa$ term orthogonal projection
\citep{donoho06:l1issparsest}.
Undoing the shrinkage after selection has an impact on the
bias-variance balance in the KL divergence or in the prediction error,
but also on the closeness-complexity balance in the estimation of the error
by an information criterion.
Indeed, the shrinkage bias occurs mainly at large values of $\lambda$, while
the variance from the false positives is mainly seen with small values of
$\lambda$. Undoing the shrinkage takes away the shrinkage bias and increases
the impact of false positives on the variance, thus shifting the optimal
bias-variance balance towards larger values of $\lambda$, meaning smaller
models with less variance.
Without shrinkage, finding the optimal model size is a more delicate task,
because the impact of false positive selections on the variance is no longer
tempered by shrinkage. Therefore, a given overestimation of the optimal
model size introduces more variance. On the other hand, taking the model too
small introduces more bias when there is no shrinkage, because the optimum
lies at smaller models, where false negatives occur more frequently.

As for the closeness-complexity balance in the information criterion, a quite
remarkable result states that for a linear regression model with normal noise,
the number of degrees of freedom of the lasso equals the size of the selected
model \citep{zou07:doflasso,tibshirani12:doflasso}.
It also equals the number of degrees of freedom in a least squares estimation
on a fixed model. In other words, with normal errors, the shrinkage bias
compensates exactly for the influence of the errors on the optimisation
process. This explains why Mallows's $C_p$ for orthogonal projection on a
given model has the same form as Stein's Unbiased Risk estimator in soft
thresholding \citep{stein81:SURE}. After undoing the shrinkage, the estimation
of $\lambda$ with minimum prediction error or KL divergence requires a tailored
information criterion. This is developed in Section \ref{sec:mirror}.

The remainder of the article is organised as follows.
Section \ref{sec:mirror} develops the idea to finetune a shrinkage based sparse
selection method (such as lasso) for minimum prediction error when using the
selection for orthogonal projection without shrinkage.
Section \ref{sec:tree} applies the methodology to subtree selection, in
particular to regression trees,
Section \ref{sec:graph} applies the methodology to graphical models
representing large multivariate normal random variables.

\section{Information criteria for use in parameter selection without shrinkage}
\label{sec:mirror}

Consider the full linear regression model
\begin{equation}
\vec{Y} = \vec{\mu} + \sigma\vec{Z} = \mathbf{X}\vec{\beta}+ \sigma\vec{Z},
\label{eq:linmodel}
\end{equation}
where $\vec{X}$ is standardised i.i.d.~noise, $\vec{\beta}$ is a sparse $m$
dimensional vector, and $\vec{Y}$ is a response vector with sample size $n$.
The statement (\ref{eq:linmodel}) assumes that the DGP is a submodel of the
full model.
Alternatively, the model in (\ref{eq:linmodel}) can be considered as a family
of approximations to the DGP, from which the member closest to the DGP is
defined the least false model.
In this definition, the distance between the DGP and the least false model is
measured, for instance, by the Kullback-Leibler divergence.

Since in the high-dimensional case $m$ may be larger than $n$, or otherwise,
the design matrix may be fully or nearly singular due to collinearity, an
estimator of $\vec{\beta}$ is searched for in a two steps regularisation
procedure. The first step computes a pilot estimator
$\widecheck{\vec{\beta}}_\lambda$, using a relatively fast selection and
estimation procedure. The prototype of a pilot procedure is the lasso,
where the estimator is defined by solving the $\ell_1$ regularised least
squares problem
\begin{equation}
\min_{\vec{\beta}} \norm{\vec{Y}-\mathbf{X}\vec{\beta}} + \lambda
\norm{\vec{\beta}}_1,
\end{equation}
where
\(
\norm{\vec{\beta}}_1 = \sum_{j=1}^m |\beta_j|.
\)
Solving the $\ell_1$ regularised least squares problem leads to sample
dependent size $\widehat{\kappa}_\lambda$. As an alternative, regularisation
can be achieved by fixing the cardinality of the selection, say $\kappa$, and
from there find the $\widehat{\lambda}_\kappa$ that leads to the best selection
$\mathrm{S}_\kappa$ with cardinality $\kappa$. The subsequent discussion will
therefore consider all quantities as function of $\kappa$ or indexed by
$\kappa$ instead of $\lambda$.
Finetuning by $\kappa$ instead of $\lambda$ is particularly interesting in
selection procedures, other than lasso, that do not rely on an
explicit regularisation parameter. An example is developed in
Section~\ref{sec:tree}.

The final estimator is then found as
\(
\widehat{\vec{\beta}}_\kappa = \widetilde{\mathbf{X}}_{\mathrm{S}_\kappa}
\vec{Y},
\)
where $\widetilde{\mathbf{X}}_{\mathrm{S}}$ is referred to as the
analysis matrix, influence matrix or hat matrix, associated with the selection
$\mathrm{S}$ (for the sake of simplicity in the notations, the subscript
$\kappa$ is omitted in expressions holding for any selection $\mathrm{S}$).
A typical choice of the hat matrix $\widetilde{\mathbf{X}}_{\mathrm{S}}$,
for a selection $\mathrm{S}$ is given by the least squares solution
\(
\widetilde{\mathbf{X}}_{\mathrm{S}} = 
(\mathbf{X}_{\mathrm{S}}^T\mathbf{X}_{\mathrm{S}})^{-1}
\mathbf{X}_{\mathrm{S}}^T
\)
where $\mathbf{X}_{\mathrm{S}}$ is the submatrix of $\mathbf{X}$ containing all
columns $j \in \mathrm{S}$. 
\begin{assump}(Projection)
It is assumed in this article that
\(
\mathbf{P}_{\mathrm{S}} =
\mathbf{X}_{\mathrm{S}}\widetilde{\mathbf{X}}_{\mathrm{S}}
\)
is a projection, i.e., an idempotent matrix. The projection is not necessarily
an orthogonal projection, i.e., $\mathbf{P}_{\mathrm{S}}$ is not necessarily
symmetric.
\end{assump}

The procedure includes a finetuning of the regularisation parameter $\kappa$,
for which the objective is to minimise the PE of the outcome,
\[
\mathrm{PE}(\widehat{\vec{\beta}}_\kappa)
= {1 \over n}
E\norm{\mathbf{X}\vec{\beta}-\mathbf{X}\widehat{\vec{\beta}}_\kappa}_2^2.
\]
The PE is estimated unbiasedly by the non-studentised version of Mallows's
$C_p$,
\begin{equation}
\Lambda(\widehat{\vec{\beta}}_\kappa)
= {1 \over n} \mathrm{SS_E}(\widehat{\vec{\beta}}_\kappa) 
               + {2\nu_\kappa \over n} \sigma^2 - \sigma^2,
\label{eq:defCpnonstud}
\end{equation}
where
\(
\mathrm{SS_E}(\widehat{\vec{\beta}}_\kappa) = \norm{\vec{e}_\kappa}_2^2,
\)
with
\(
\vec{e}_\kappa
=
\vec{Y}-\widehat{\vec{\mu}}_\kappa
=
\vec{Y}-\mathbf{X}\widehat{\vec{\beta}}_\kappa
\)
the residual vector, and where
$\nu_\kappa$ are the generalised degrees of freedom
\citep{ye98:modelselection}, defined and developed by
\[
\nu_\kappa
=
{1 \over \sigma^2}
E\left[\vec{\varepsilon}^T(\vec{\varepsilon}-\vec{e}_\kappa)\right]
=
{1 \over \sigma^2}
E\left[\vec{\varepsilon}^T
       (\vec{Y}-\vec{\mu}-\vec{Y}+\widehat{\vec{\mu}}_\kappa)\right]
=
{1 \over \sigma^2}
E\left[\vec{\varepsilon}^T\widehat{\vec{\mu}}_\kappa\right]
\]
If the nuisance parameter $\sigma^2$ is not available or hard to estimate
independently from the parameter selection process, then minimisation of
(\ref{eq:defCpnonstud}) can be replaced with a generalised cross validation
\citep{jansen15:gcv}
\begin{equation}
\mathrm{GCV}(\widehat{\vec{\beta}}_\kappa)
=
{\displaystyle
{1 \over n} \mathrm{SS_E}(\widehat{\vec{\beta}}_\kappa)
\over\displaystyle
\left(1-{\nu_\kappa \over n}\right)^2}.
\label{eq:defGCV}
\end{equation}
If the final estimator were taken to be just the pilot estimator (including the
lasso shrinkage, that is), 
$\widehat{\vec{\beta}}_\kappa = \widecheck{\vec{\beta}}_\kappa$, 
then the degrees of freedom would be simply $\nu_\kappa = \kappa$,
\citep{zou07:doflasso,tibshirani12:doflasso}.
While the pilot estimator suffers from the above mentioned tolerance
of false positives, the minimum GCV shrinkage estimator
$\widecheck{\vec{\beta}}_{\kappa^*}$ can be used to define an estimator of the
variance
\[
\widehat{\sigma}^2 = {1 \over \nu_{\kappa^*}}
\mathrm{SS_E}\left(\widecheck{\vec{\beta}}_{\kappa^*}\right)
=
{1 \over \kappa^*}
\mathrm{SS_E}\left(\widecheck{\vec{\beta}}_{\kappa^*}\right)
\]
for use in the finetuning of the second step of the selection
procedure, which is the step leading from the pilot
$\widecheck{\vec{\beta}}_\kappa$ to the final $\widehat{\vec{\beta}}_\kappa$.
The finetuning in that second step
aims at minimising the prediction error or its estimator
(\ref{eq:defCpnonstud}), this time with taking $\nu_\kappa$ to be the degrees
of freedom under orthogonal projection without shrinkage.

%

The degrees of freedom are further developed as
\[
\nu_\kappa
=
{1 \over \sigma^2}
E\left[\vec{\varepsilon}^T\mathbf{P}_{\mathrm{S}_\kappa} \vec{Y}\right]
=
{1 \over \sigma^2}
E\left[\sigma\vec{Z}^T\mathbf{P}_{\mathrm{S}_\kappa}
        (\vec{\mu}+\sigma\vec{Z})\right]
=
E\left[\norm{\mathbf{P}_{\mathrm{S}_\kappa} \vec{Z}}^2_2\right] 
+
{1 \over \sigma}\vec{\mu}^T E\left[\mathbf{P}_{\mathrm{S}_\kappa} \vec{Z}\right].
\]
If the selection $\mathrm{S}_\kappa$ were independent from the sample, then
$E\left[\norm{\mathbf{P}_{\mathrm{S}_\kappa} \vec{Z}}^2_2\right]$ would be
equal to $\mathrm{Tr}(\mathbf{P}_{\mathrm{S}_\kappa}) = \kappa$,
and $E\left[\mathbf{P}_{\mathrm{S}_\kappa} \vec{Z}\right]$
would be the zero vector. The interaction between the noise and the selection
makes the set $\mathrm{S}_\kappa$ and the vector $\vec{Z}$ depend from each
other, which is precisely the topic of this paper.
\begin{assump}
As in \citet{jansen14:ICsparsity}, it is assumed here that the dependence of
$\mathrm{S}_\kappa$ and $\vec{Z}$ has an effect on the magnitudes of the errors
after selection, and not so much on the signs. More precisely, it is assumed
that
\begin{equation}
\nu_\kappa
=
E\left[\norm{\mathbf{P}_{\mathrm{S}_\kappa} \vec{Z}}^2_2\right]
+ o\left[\mathrm{PE}(\widehat{\vec{\beta}}_\kappa)\right] 
\mathrm{~as~} n \rightarrow \infty.
\label{eq:dofHT}
\end{equation}
\end{assump}

The offset
\begin{equation}
m_\kappa = (\nu_\kappa-\kappa)\sigma^2/n
= E\left[\norm{\mathbf{P}_{\mathrm{S}_\kappa} \vec{Z}}^2_2 - \kappa\right] 
\sigma^2/n
+ o\left[\mathrm{PE}(\widehat{\vec{\beta}}_\kappa)\right]
\label{eq:defmirror}
\end{equation}
corrects a Mallows's $C_p$ criterion assessing a given, \emph{fixed} model
\(
\Delta(\widehat{\vec{\beta}}_\kappa)
= {1 \over n} \mathrm{SS_E}(\widehat{\vec{\beta}}_\kappa) 
               + {2\kappa \over n} \sigma^2 - \sigma^2,
\)
for use in parameter selection, as indeed, from (\ref{eq:dofHT}),
\(
E(\Lambda(\widehat{\vec{\beta}}_\kappa)) =
E(\Delta(\widehat{\vec{\beta}}_\kappa)) +
2m_\kappa + o\left[\mathrm{PE}(\widehat{\vec{\beta}}_\kappa)\right],
\)
keeping in mind that
$E\left(\Lambda(\widehat{\vec{\beta}}_\kappa)\right) =
\mathrm{PE}(\widehat{\vec{\beta}}_\kappa)$.
The offset $m_\kappa$ thus describes the effect of the selection procedure
onto the degrees of freedom.
Furthermore, the correction can be seen as a double reflection
\citep{jansen14:ICsparsity}, on both sides of a
``mirror'' function $\mathrm{PE}(\widehat{\vec{\beta}}_{\mathrm{O}_\kappa})$, 
in the sense that
\begin{equation}
m_\kappa =
\mathrm{PE}(\widehat{\vec{\beta}}_\kappa)-
\mathrm{PE}(\widehat{\vec{\beta}}_{\mathrm{O}_\kappa})
+ o\left[\mathrm{PE}(\widehat{\vec{\beta}}_\kappa)\right]
= 
\mathrm{PE}(\widehat{\vec{\beta}}_{\mathrm{O}_\kappa})-
E\left(\Delta(\widehat{\vec{\beta}}_\kappa)\right)
+  o\left[\mathrm{PE}(\widehat{\vec{\beta}}_\kappa)\right].
\end{equation}
The mirror function $\mathrm{PE}(\widehat{\vec{\beta}}_{\mathrm{O}_\kappa})$ is
given by the prediction error of a least squares estimator
\(
\widehat{\vec{\beta}}_{\mathrm{O}_\kappa}
=
\left(\mathbf{X}_{\mathrm{O}_\kappa}^T\mathbf{X}_{\mathrm{O}_\kappa}\right)^{-1}
\mathbf{X}_{\mathrm{O}_\kappa}^T\vec{Y},
\)
on a selection $\mathrm{O}_\kappa$ made by an oracle observing the response
without noise, $\mathbf{X}\vec{\beta}$.


The evaluation of the mirror correction (\ref{eq:defmirror}) in practical
situations, where no oracle is available, is non-trivial, as it depends on the
adopted selection procedure, the size and structure (nested, trees, etc.)
of the set of models under consideration, the design matrix $\mathbf{X}$.
A bootstrap or any other resampling procedure, for instance, is hard to
set up, precisely because $m_\kappa$ describes the interaction between the noise
in the original sample and the selection.
Monte-Carlo simulations can be used in the calculation of the subsequent
estimation.
The estimation developed in this paper starts from the straightforward
application of the law of iterated expectations, stating that
$m_\kappa = E(\widehat{m}_\kappa)$ where
\begin{equation}
\widehat{m}_\kappa = \left[E\left(\norm{\mathbf{P}_{\mathrm{S}_\kappa}
\vec{Z}}^2_2\middle|\mathrm{S}_\kappa\right) - \kappa\right] \sigma^2/n,
\label{eq:EnormPSgivenS}
\end{equation}
which is random, sample dependent through the conditioning on the sample
dependent selection $\mathrm{S}_\kappa$.
The central point in the development in the subsequent sections (in particular
Section \ref{subsec:dofgivenS})
will be the understanding of what it means to have observed the event
$\mathrm{S}_\kappa$. 
In other words, we need to quantify the exact information provided by the
selection $\mathrm{S}_\kappa$ at the level of each selected or active
parameters $\beta_l$, $l \in \mathrm{S}_\kappa$. 
The issue is related to problems in the domain of post-selection inference.
Post-selection inference may proceed in basically two ways. The first,
generalistic approach aims at confidence intervals that are valid, regardless
of the selection procedure (possibly limited to a certain class of selection or
models) \citep{berk13:validpostselectioninference}. The second approach,
related to the discussion in this paper, operates on a conditional, rather
than a generalistic level, looking for the conditional distribution of
estimators, given a specific selection procedure
\citep{charki18:AICpostselection}.
Our paper studies the effect of the selection on the degrees of
freedom and the information criterion, not on the subsequent inference.

The remainder of this paper is devoted to the development and the estimation
of the correction $m_\kappa$ in two graphical models.

\section{Finetuning a tree selection}
\label{sec:tree}

The first application of the proposed selection based information criterion
consists in finetuning a backtracking algorithm for best $\kappa$-subtree
selection.
Applications are situated in Best Orthogonal Basis selection
\citep{coi-wic92:entropy},
classification and regression trees \citep{breiman84:CART},
tree structured wavelet regression \citep{jansen22:waveletbook},
and wavelet packets pruning.

\subsection{A tree structured model of covariates}

The best $\kappa$-subtree selection operates on the linear regression model in
(\ref{eq:linmodel}). It replaces a lasso procedure, such as
LARS \citep{efron04:lars} or a proximal gradient or subgradient method.
The tree structured selection rests on two assumptions, leading to a more
specialised selection procedure. 
The first assumption concerns the set of models to choose from.
Denoting a model by the subset of the indices
$\mathrm{S} \subset \{1,2,\ldots,m\}$ corresponding to the nonzero components
of the parameter vector $\vec{\beta}$, the tree structured selection restricts 
the search to subsets satisfying an imposed hierarchy in a way explained below.
The hierarchy is supposed to reflect additional information on the physical
nature of the covariates, by stating that a given covariate cannot be part of a
model unless at least another specific covariate is selected as well. 
It thus defines for every component $i\in \{1,2,\ldots,m\}$, termed a node in
this context, a unique parent node $p(i) \in \{0,1,\ldots,m\}$, where
$p(i)=0$ means that the node has no parent, i.e., it is a root.
An important special case is that of the binary tree rooted at node $1$, which
has $p(i) = \floor{i/2}$, where $\floor{x}$ is the floor function of $x$.
The tree selection developed below works on general, not necessarily binary, 
trees and even in the presence of multiple roots. An extreme example of the
latter is the case where $p(i)=0$ for all nodes $i$, leading to the situation 
where there is no hierarchy and all nodes are roots.
\begin{assump}(tree structured selection)
A valid selection $\mathrm{S}$ satisfies the hierarchy in the sense that
$i \in \mathrm{S} \Rightarrow p(i) \in \mathrm{S}$.
\end{assump}
The second assumption puts restrictions on collinearity.
\begin{assump}(frame condition)
Let the $m$ columns of the design matrix be $\ell_2$ normalised, i.e., the
diagonal of $\mathbf{X}^T\mathbf{X}$ is the identity matrix. Then we assume the
existence of a matrix $\widetilde{\mathbf{X}}^T$ with the same size 
$n \times m$  as that of $\mathbf{X}$, and of two positive constants
$\gamma$ and $\Gamma$, independent from $n$ and $m$ so that for any
$\vec{\mu}\in\RR^n$,
\[
{\gamma \over m} \norm{\widetilde{\mathbf{X}}\vec{\mu}}_2^2
\leq
{1 \over n} \norm{\vec{\mu}}_2^2
\leq
{\Gamma \over m} \norm{\widetilde{\mathbf{X}}\vec{\mu}}_2^2,
\]
while $\mathbf{X}\widetilde{\mathbf{X}}$ is $m/n$ times the $n \times n$
identity matrix.
For each selection $\mathrm{S}$, the estimation is supposed to be constructed
by composing the hat matrix $\widetilde{\mathbf{X}}_{\mathrm{S}}$ from the
columns of $\widetilde{\mathbf{X}}$ corresponding to the elements in
$\mathrm{S}$.
\label{assump:frame}
\end{assump}
In this setting, the hat matrix $\widetilde{\mathbf{X}}_{\mathrm{S}}$
is not constructed after but \emph{before} selection, from the
full hat matrix $\widetilde{\mathbf{X}}$, whose columns depend
on all columns of $\mathbf{X}$, not only on the selected ones in
$\mathbf{X}_{\mathrm{S}}$. More precisely, we have
\(
\widetilde{\mathbf{X}}_{\mathrm{S}} = (n/m)\mathbf{D}_{\mathrm{S}}
\widetilde{\mathbf{X}}.
\)
With $k$ the cardinality of $\mathrm{S}$, the matrix $\mathbf{D}_{\mathrm{S}}$
is a $k \times n$ diagonal selection with elements
$D_{\mathrm{S};ij} = 1$ if the $i$th element of the sorted sequence from
$\mathrm{S}$ equals $j$, for $i=1,2,\ldots,k$.

An example of the matrix $\widetilde{\mathbf{X}}$ is the pseudo-inverse
(or Moore-Penrose inverse), while the smallest and largest singular values of
$\mathbf{X}$ can be associated to $\sqrt{n\gamma/m}$ and $\sqrt{n\Gamma/m}$
respectively. 
When $n=m$, such as in a fast (decimated) wavelet transform, the matrix
$\widetilde{\mathbf{X}}$ corresponds to the forward (analysis) data
transform while the matrix $\mathbf{X}$ represents the reconstruction.
In non-orthogonal transforms, the forward transform $\widetilde{\mathbf{X}}$
does not coincide with the pseudo-inverse of the reconstruction.

The vector $\vec{\beta} = (n/m)\widetilde{\mathbf{X}}\vec{\mu}$ is a valid
model, in the sense that $\mathbf{X}\vec{\beta}$ predicts the observations
from the DGP exactly.
In the presence of noise, $\widetilde{\vec{Y}} =
(n/m)\widetilde{\mathbf{X}}\vec{Y}$ is an unbiased vector of
pseudo-observations of $\vec{\beta}$.
The sparse estimator $\widehat{\vec{\beta}}_\mathrm{S}$ is then given by
diagonal selection of the pseudo-observations,
\(
\widehat{\vec{\beta}}_\mathrm{S} = \mathbf{D}_{\mathrm{S}}\widetilde{\vec{Y}}.
\)

Because of Assumption \ref{assump:frame}, the prediction error of an estimator
$\widehat{\vec{\beta}}_\mathrm{S}$ is bounded by
\[
{\gamma m^2\over n^2}\mathrm{R}(\widehat{\vec{\beta}}_\mathrm{S})
\leq \mathrm{PE}(\widehat{\vec{\beta}}_\mathrm{S}) \leq
{\Gamma m^2\over n^2}\mathrm{R}(\widehat{\vec{\beta}}_\mathrm{S}),
\]
where $\mathrm{R}(\widehat{\vec{\beta}}_\mathrm{S})$ is the risk or prediction
error on the selected pseudo-observations,
\begin{equation}
\mathrm{R}(\widehat{\vec{\beta}}_\mathrm{S}) = {1 \over m}
E\norm{\widehat{\vec{\beta}}_\mathrm{S}-\vec{\beta}}_2^2
\label{eq:defR}
\end{equation}

The prediction error $\mathrm{R}(\widehat{\vec{\beta}}_\mathrm{S})$ can be
estimated by an information criterion as in (\ref{eq:defCpnonstud}),
defined on the pseudo-observations. We have
\begin{equation}
\mathrm{SS_E}(\widehat{\vec{\beta}}_\mathrm{S})
=
\norm{\widetilde{\vec{Y}}-
\mathbf{D}_{\mathrm{S}}\widetilde{\vec{Y}}}_2^2
=
\norm{\widetilde{\vec{Y}}}-\sum_{\ell \in \mathrm{S}}
\widetilde{Y}_{\ell}^2.
\label{eq:SSEtree}
\end{equation}
With $\widetilde{\vec{Z}} = \widetilde{\mathbf{X}}\vec{Z}$,
the degrees of freedom are given by
\(
\nu_\mathrm{S}
=
E\left[\norm{\mathbf{D}_{\mathrm{S}}\widetilde{\vec{Z}}}_2^2\right]
=
E\left[\sum_{\ell\in\mathrm{S}} \widetilde{Z}_\ell^2\right],
\)
taking a possible sample dependence of the selection $\mathrm{S}$ into account.

\subsection{Best $\kappa$-subtree selection}

The best $\kappa$-term model $\mathrm{S}_\kappa$ is supposed to be the subtree
that maximises the amount of information accumulated in its $\kappa$ elements,
under the constraint of the hierarchy imposed by the parent function $p(i)$.
The amount of information is quantified by an accumulative mass function
$M(\mathrm{S};\widetilde{\vec{Y}})$, where
\begin{equation}
M(\mathrm{S};\vec{x}) = \sum_{i \in \mathrm{S}} M_i(x_i).
\label{eq:defTq}
\end{equation}
The mass function is accumulative in the sense that if $\mathrm{S}_1 \cap
\mathrm{S}_2 = \emptyset$, then
\(
M(\mathrm{S}_1\cup\mathrm{S}_2;\vec{x}) =
M(\mathrm{S}_1;\vec{x})+ M(\mathrm{S}_2;\vec{x}).
\)
The elementary mass functions $M_i(x_i)$ are supposed to be convex with
absolute minimum at zero. Most often these functions will be taken to be
symmetric, leading to non-decreasing function of $|x_i|$. Simple examples
include the choices $M_i(x_i) = |x_i|$ or $M_i(x_i) = |x_i|^2$, the latter
being the default choice in the subsequent discussion. The function $M_i$
may depend on $i$, allowing us to attach more importance to some nodes,
regardless of their values. For instance, nodes close to the root may be
a priori more important that nodes further away.

The backtracking algorithm for searching $\mathrm{S}_\kappa$ finds the
solutions for all $\kappa$ at once. It extends weakest link or cost complexity
pruning techniques to trees whose subsequent best $\kappa$-subtrees
(for increasing $\kappa$, that is) cannot be guaranteed to be nested.
It proceeds as follows
\begin{enumerate}
\item
Take as input a vector $\vec{x}$ in $n$ nodes on which lies a hierarchy,
defined by a function $p(i)$ that maps each node onto its parent. The objective
is to find for each $\kappa$ the selection $\mathrm{S}_\kappa$ maximising
$M(\mathrm{S};\vec{x})$, defined in (\ref{eq:defTq}) among all subsets
$\mathrm{S}$ with $\kappa$ elements and satisfying the hierarchy.
\item
\textbf{If} the set of nodes $i$ with $p(i)=0$ has more than one element, then
introduce a ``superroot'' as a parent to all these nodes. The result of this
step is a rooted tree. Define for each node $i$, including superroot $i=0$, the
set of its children $\mbox{C}_i = \{j|p(j) = i\}$.
Let $\mbox{A}$ denote the set of active nodes, i.e., nodes open for further
processing, initialising $\mbox{A} \leftarrow \{1,2,\ldots,n\}$.
Initialise $j \leftarrow 0$, and the depth $d \leftarrow 0$.
\item
\textbf{While} the set $\mbox{A}$ is not empty,
\begin{enumerate}
\item
First \emph{descend} into the tree, as far as possible\\
\textbf{While} $\mbox{C}_j \cap \mbox{A}$ is not empty:
\begin{itemize}
\item
Set $j \leftarrow \min(\mbox{C}_j \cap \mbox{A})$.
\item
Set $d \leftarrow d+1$.
\item
Initialise or re-initialise the best $1$ subtree, rooted at the current node at
depth $d$,\\
$\mathrm{S}_{1,d} \leftarrow \{j\}$.
\end{itemize}
\item
We are now at a node $j$ that has no active children. This node is
\emph{merged} with its parent, $i = p(j)$.
Find the best combinations of subtrees of parent and child, thus
incorporating the subtrees of $j$ into those of $i$.
\\
\textbf{For} $\kappa=1,2,\ldots$
\begin{itemize}
\item
Find 
\(\displaystyle
\ell = \arg\max_{l=1,2,\ldots,\kappa}
\left[M(\mathrm{S}_{l,d-1};\vec{x})+
      M(\mathrm{S}_{\kappa-l,d};\vec{x})\right].
\)
\item
Set
\(
\mathrm{S}_{\kappa,d-1} = \mathrm{S}_{\ell,d-1} \cup \mathrm{S}_{\kappa-\ell,d}.
\)
\end{itemize}
\item
Take $j$ out of the active set $\mbox{A}$. Then go one up one level, i.e.,
$d \leftarrow d-1$, and $j \leftarrow i$. 
\\
If the new $j$ still has children within the active set, a new descent will
take place along one of these children. Otherwise, $j$ will be merged with its
parent, and so, until all nodes have been visited and merged.
\end{enumerate}
\item
Set $\mathrm{S}_\kappa \leftarrow \mathrm{S}_{\kappa,0}$.
\end{enumerate}

Once the selections $\mathrm{S}_\kappa$ have been found, finetuning amounts to
choosing $\kappa$ according to the criterion obtained by substitution of
(\ref{eq:SSEtree}) into (\ref{eq:defCpnonstud}). The calculation of the degrees
of freedom in (\ref{eq:defCpnonstud}) requires some further attention, as
discussed Section \ref{subsec:dofgivenS}.

\subsection{The effect of the selection on the degrees of freedom}
\label{subsec:dofgivenS}

Further development of $\widehat{m}_\kappa$ in (\ref{eq:EnormPSgivenS}) is
based on 
symmetric mass functions $M_i(u)$. 
If $M_i(u)$ is symmetric, then there exists a random threshold
$\widehat{\theta}_{\kappa;i}$, depending on all $\widetilde{Y}_j$ with
$j\neq i$, so that
\[
i \in \mathrm{S}_\kappa \Leftrightarrow |\widetilde{Y}_i| \geq
\widehat{\theta}_{\kappa;i}.
\]
From this it follows that
\[
\widehat{m}_\kappa = (\sigma^2/n)\sum_{i \in \mathrm{S}_\kappa}
\left\{E\left[\widetilde{Z}_i^2\middle|\widetilde{Y}_i^2 >
\widehat{\theta}_{\kappa;i}^2\right]-1\right\},
\]
which, according to the main result in \citet{marquis22:group}, can be well
approximated by replacing the sparse vector $\vec{\beta}$ in the model of the
pseudo-observations $\widetilde{\vec{Y}}$ by the zero vector.
More precisely, it holds that
\begin{equation}
E\left[\widetilde{Z}_i^2\middle|\widetilde{Y}_i^2 >
\widehat{\theta}_{\kappa;i}^2\right]
=
E\left[\widetilde{Z}_i^2\middle|\widetilde{Z}_i^2 >
\widehat{\theta}_{\kappa;i}^2\right] + o(\mathrm{R}(\kappa)).
\label{eq:apprxbeta0}
\end{equation}
Unfortunately, to the best of the author's knowledge, there seems to be no fast
way to find or even approximate the functions that map the pseudo-observations
$\widetilde{\vec{Y}}$ onto the thresholds $\widehat{\theta}_{\kappa;i}$.


As an alternative, the expected values $E(\widehat{m}_\kappa)$ can be
approximated numerically quite well by running Monte Carlo simulations with
$\vec{\beta} = \vec{0}$ and computer generated, pseudo-random vectors
$\vec{Z}$ on top of the tree structure specified by the parent function $p(i)$.
The numerical simulation thus adopts the same approximation as
(\ref{eq:apprxbeta0}).

\subsection{Application to regression trees}

Let $\vec{Y} = \vec{\mu}+\vec{\eta}$, where $\vec{\eta}$ represents
uncorrelated, zero mean noise, while the vector $\vec{\mu}$ is modelled to
consist of constant segments, separated by change points from a set
$\mathrm{CP} \subset \{i+1/2,i=1,2,\ldots,n\}$, meaning
that $\mu_i = \mu_{i+1}$ unless $i+1/2 \in \mathrm{CP}$.
The objective is to identify the set $\mathrm{CP}$.
With $\widehat{\mathrm{CP}}$ the estimated set of change points, the elements
of the vector $\vec{\mu}$ can be estimated by
\[
\widehat{\mu}_i = \overline{Y}_{I(i)} = {1 \over |I(i)|} \sum_{\ell \in I(i)}
Y_\ell,
\]
where $I(i)$ is the set $\{l,l+1,\ldots,r\}$ so that
$i \in I(i)$, and
$l-1/2,r+1/2 \in \widehat{\mathrm{CP}}$, while
$\{\ell+1/2,\ell\in I(i)\} \cap \widehat{\mathrm{CP}} = \emptyset$.

The search for candidate change points proceeds through a greedy tree
construction, starting with $I_{0,0} = \{1,2,\ldots,n\}$ and $n_0 = 1$. Then
for $j=0,1,2,\ldots$ and for $\ell = 0,\ldots,n_j-1$, find a partitioning
of the set $I_{j,\ell} = I_{j+1,2\ell} \cup I_{j+1,2\ell+1}$, as long as
$I_{j,\ell}$ has at least two elements. The partitioning is defined by a new
candidate change point $t_{j,\ell}+1/2$, fixing
$I_{j+1,2\ell} = \{i \in I_{j,\ell};i<t_{j,\ell}+1/2\}$ and $I_{j+1,2\ell+1}$
as its complement. The value of $t_{j,\ell}$ is chosen to maximise a contrast
function $c_{j,\ell} = c(\vec{Y},I_{j,\ell},I_{j+1,2\ell})$, in which the
argument $I_{j+1,2\ell}$ depends on $t_{j,\ell}$. It is clear that the set of
all $I_{j,\ell}$ constitute a binary tree rooted at $I_{0,0}$, referred to as
the refinement tree. When the refinement is pursued until all leaves
$I_{j,\ell}$ are singletons, then the tree has $2n-1$ nodes in total, root and
leaves included.

The contrast $c_{j,\ell}$ may or may not coincide with the absolute or squared
value of the offset or detail
\[
d_{j,\ell} = {\overline{Y}_{j+1,2\ell+1} - \overline{Y}_{j+1,2\ell} \over
\sqrt{{1\over n_{j+1,2\ell+1}} + {1\over n_{j+1,2\ell}}}},
\]
where $n_{j,\ell}$ is the cardinality of $I_{j,\ell}$ and
$\overline{Y}_{j,\ell}$ the average value of $\vec{Y}$ on $I_{j,\ell}$.
The $n-1$ detail coefficients $d_{j,\ell}$ can each be associated with the
corresponding internal node $I_{j,\ell}$ in the refinement tree.
The detail coefficients are completed by a single overall normalised average
value
\(
s_{0,0} = \sqrt{n}\cdot\overline{Y}.
\)
Within a given refinement tree, the mapping of the vector $\vec{Y}$ onto the
vector of details and overall average is an orthogonal transform.
This transform is known as the data-adaptive, unbalanced, orthogonal
Haar-wavelet transform (AUHT) \citep{girardi97:unbalanced}. Selection of
$\widehat{\mathrm{CP}}_\kappa$ then amounts to the selection of the best
$\kappa$-subtree $T_\kappa$ of the refinement tree.


The quality of $T_\kappa$ can be measured in the prediction
error
\(
\mathrm{PE}(\vec{\mu}_\kappa) = 
{1 \over n} E\norm{\widehat{\vec{\mu}}-\vec{\mu}}_2^2
\)
and estimated by
\(
\Lambda(\vec{\mu}_\kappa) = 
{1 \over n} \norm{\widehat{\vec{\mu}}-\vec{Y}}_2^2 + {2\nu_\kappa \over n}
\sigma^2 -\sigma^2.
\)
Thanks to the orthogonality of the AUHT, it holds that
\[
\Lambda(\vec{\mu}_\kappa) = 
{1 \over n} \sum_{(j,\ell)\not\in T_\kappa} d_{j,\ell}^2 +
{2\nu_\kappa \over n} \sigma^2 -\sigma^2.
\]
The degrees of freedom $\nu_\kappa$ can be approximated using
(\ref{eq:defmirror}), which amounts to
\(
\nu_\kappa \approx
E\left[\norm{\mathbf{P}_{\mathrm{S}_\kappa}\vec{Z}}^2_2\right].
\)
The Monte Carlo calculation proceeds by running the tree selection procedure
on a pseudo-random vector $\vec{Z}$.

\subsection{A simulation study with Poisson noise}

\begin{figure}[h!]
\begin{center}
\includegraphics[width=0.8\textwidth]{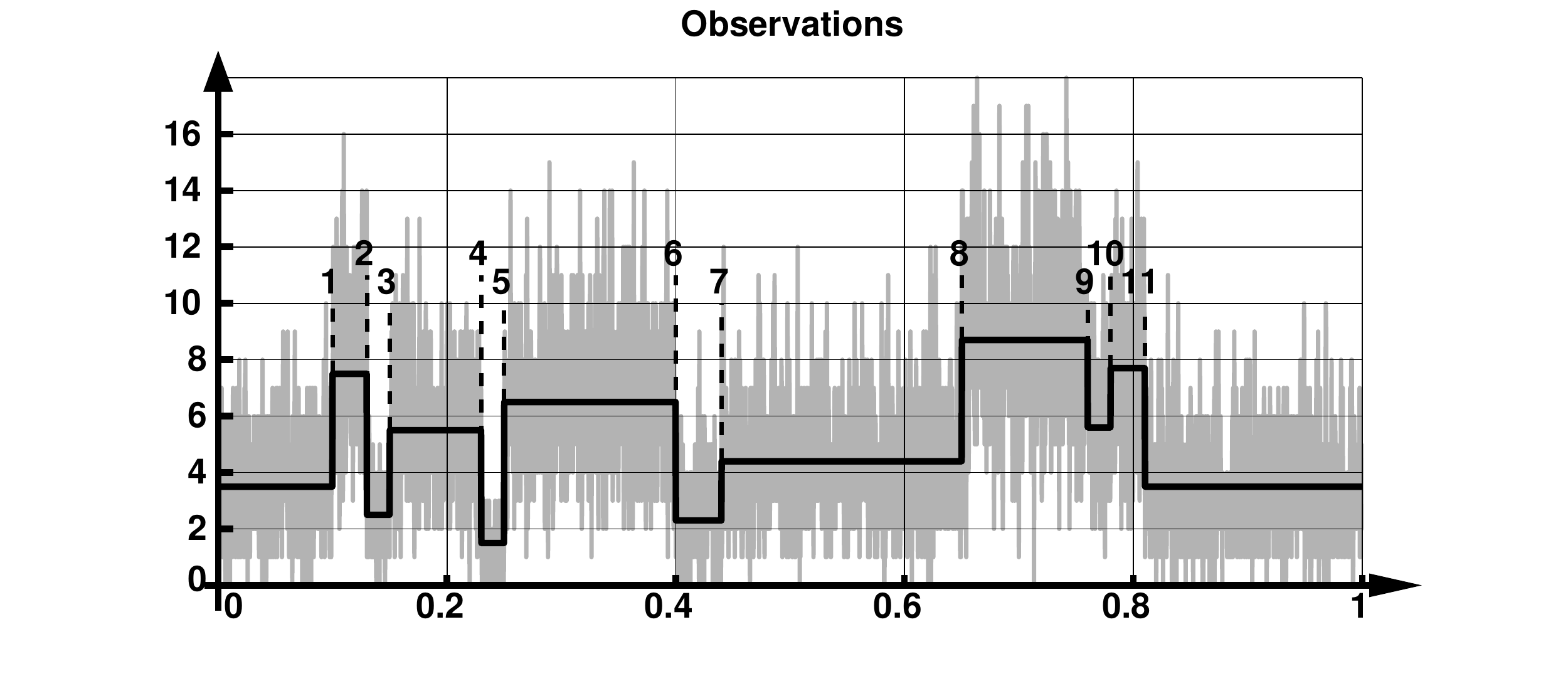}
\caption{Poisson intensity curve (black line) along with $n=4000$ corresponding
pseudo-random Poisson observations.
The objective is to retrieve the change points (discontinuities) in the black
curve.
The intensity curve is obtained by adding the value 3.5 to the values of the
well known `blocks' test function \citep{don-jon94:isaws}.
The intensity curve has $n_1=11$ change points, numbered 1 to 11 for further
reference.}
\label{fig:blockpoisson}
\end{center}
\end{figure}

If the noise in the model $\vec{Y} = \vec{\mu}+\vec{\eta}$ is uncorrelated but
heteroscedastic, then the AUHT vector $\vec{d}$ will be heteroscedastic as
well. Let $\widetilde{\mathbf{W}}$ denote the data-adaptive matrix that maps
$\vec{Y}$ onto $\vec{d}$ and define
\(
\vec{v} = \widetilde{\mathbf{W}}\vec{\mu},
\)
then a large value of $d_{j,\ell}$ may be due to large variance or to a large
absolute expected value. The distinction between these two cases is quantified
by taking a standardised prediction error,
\[
\mathrm{PE}(\widehat{\vec{v}}_\kappa) = {1 \over n} \sum_{j,\ell}
\left({\widehat{v}_{j,\ell}-v_{j,\ell} \over \sigma_{j,\ell}}\right)^2,
\]
where $\sigma_{j,\ell}^2 = \var(d_{j,\ell})$.
This prediction error is estimated by
\begin{equation}
\Lambda(\vec{v}_\kappa) = 
{1 \over n} \sum_{(j,\ell)\not\in T_\kappa}
{d_{j,\ell}^2\over\sigma_{j,\ell}^2} +
{2\nu_\kappa \over n} - 1.
\label{eq:Cptree}
\end{equation}
We apply the procedure to Poisson distributed observations with intermittent
intensities, as illustrated in Figure \ref{fig:blockpoisson}. The samplesize is
$n = 4000$.
The intensity curve, depicted in solid black line, is taken from the well
known `blocks' test function \citep{don-jon94:isaws}, vertically translated 
by adding 3.5, in order to create comparable settings as in the simulation
study in \citet{jansen07:changepoint}.

The variances $\sigma_{j,\ell}^2$ are estimated
as the diagonal elements of the estimated covariance matrix
\[
\widehat{\mathbf{\Sigma}}_{\vec{d}} =
\widetilde{\mathbf{W}}\widehat{\mathbf{\Sigma}}_{\vec{Y}}\widetilde{\mathbf{W}}^T,
\]
where the matrix $\widehat{\mathbf{\Sigma}}_{\vec{Y}}$ is obtained as the
diagonal matrix whose elements are pilot estimators $\widehat{\mu}_{0;i}$ of
$\mu_i = \var(Y_i)$. The pilot estimator is obtained from an unstructured,
threshold based selection
\(
\widehat{\vec{\mu}}_0 = \widetilde{\mathbf{W}}^{-1}\widehat{\vec{v}}_0,
\)
where
\[
\widehat{v}_{0;j,k} = \mathrm{ST}(d_{j,k}/\sigma_{00,j,k},\lambda)
\cdot\sigma_{00,j,k}.
\]
Here
\(
\mathrm{ST}(x,\lambda) = \sign(x)\cdot(|x|-\lambda)\cdot I(|x|>\lambda)
\)
is the soft-threshold function, in which $I(|x|>\lambda)$ is the indicator
function, i.e., $I(|x|>\lambda) = 1 \Leftrightarrow |x| >\lambda$ and
$I(|x|>\lambda) = 0$ otherwise. The threshold in the pilot estimator is
selected by a GCV or $C_p$ criterion as well. The pre-pilot estimator
$\sigma_{00,j,k}$ is obtained as a diagonal element in the unbiased estimator
of the covariance matrix
\(
\widetilde{\mathbf{W}}\mathrm{diag}(\vec{Y})\widetilde{\mathbf{W}}^T.
\)
The reason for not taking this unbiased estimator in the tree structured
selection and estimation lies in the large variance of the unbiased estimator,
adding fluctuations to the standardised data, thus falsely suggesting the
presence of change points.

\begin{figure}[h!]
\begin{center}
\includegraphics[width=0.8\textwidth]{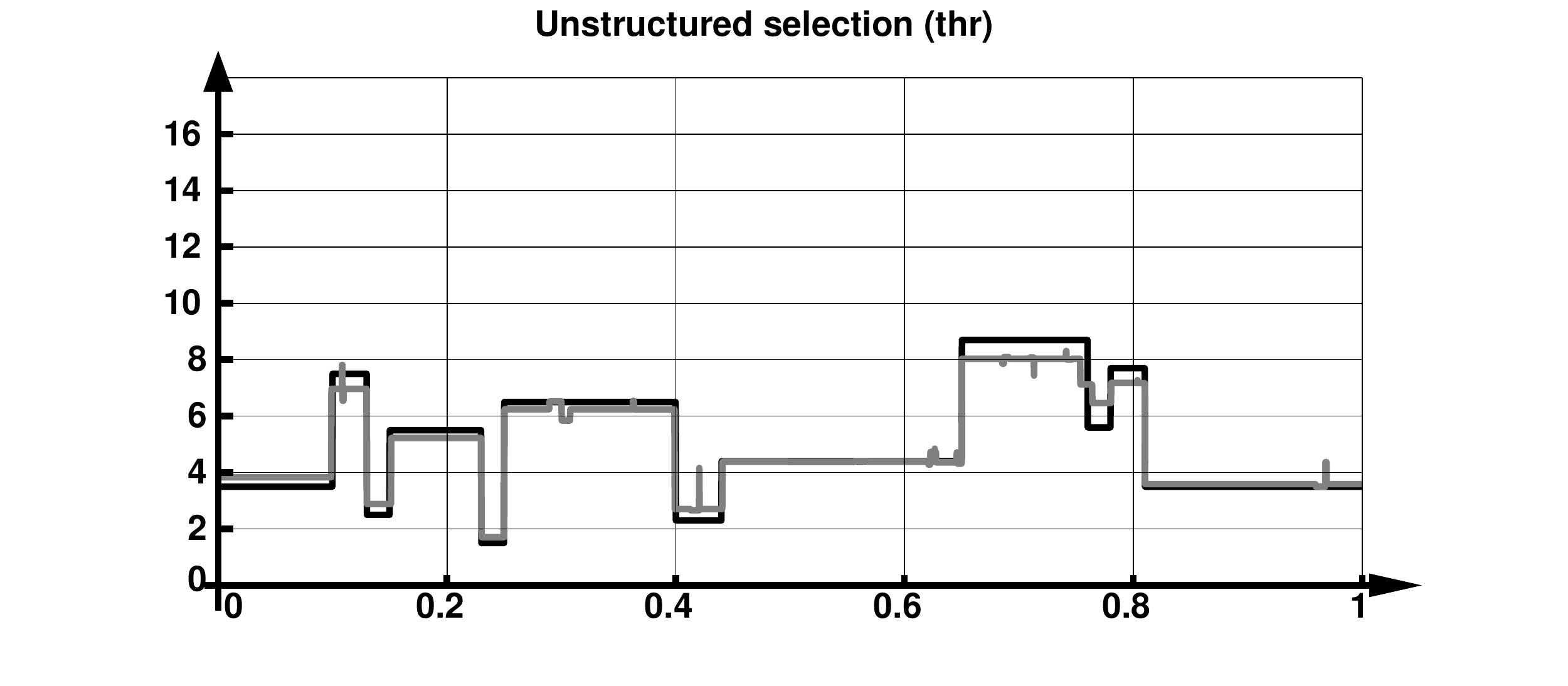}
\caption{Reconstruction of the Poisson intensity curve (black line) in Figure
\ref{fig:blockpoisson} using soft-thresholding applied to the AUHT
coefficients of the observations (grey line in Figure \ref{fig:blockpoisson}). 
The threshold is finetuned by minimisation of the Generalised
Cross Validation score, defined in (\ref{eq:defGCV}).}
\label{fig:blockpoissonthr}
\end{center}
\end{figure}

\begin{figure}[h!]
\begin{center}
\includegraphics[width=0.8\textwidth]{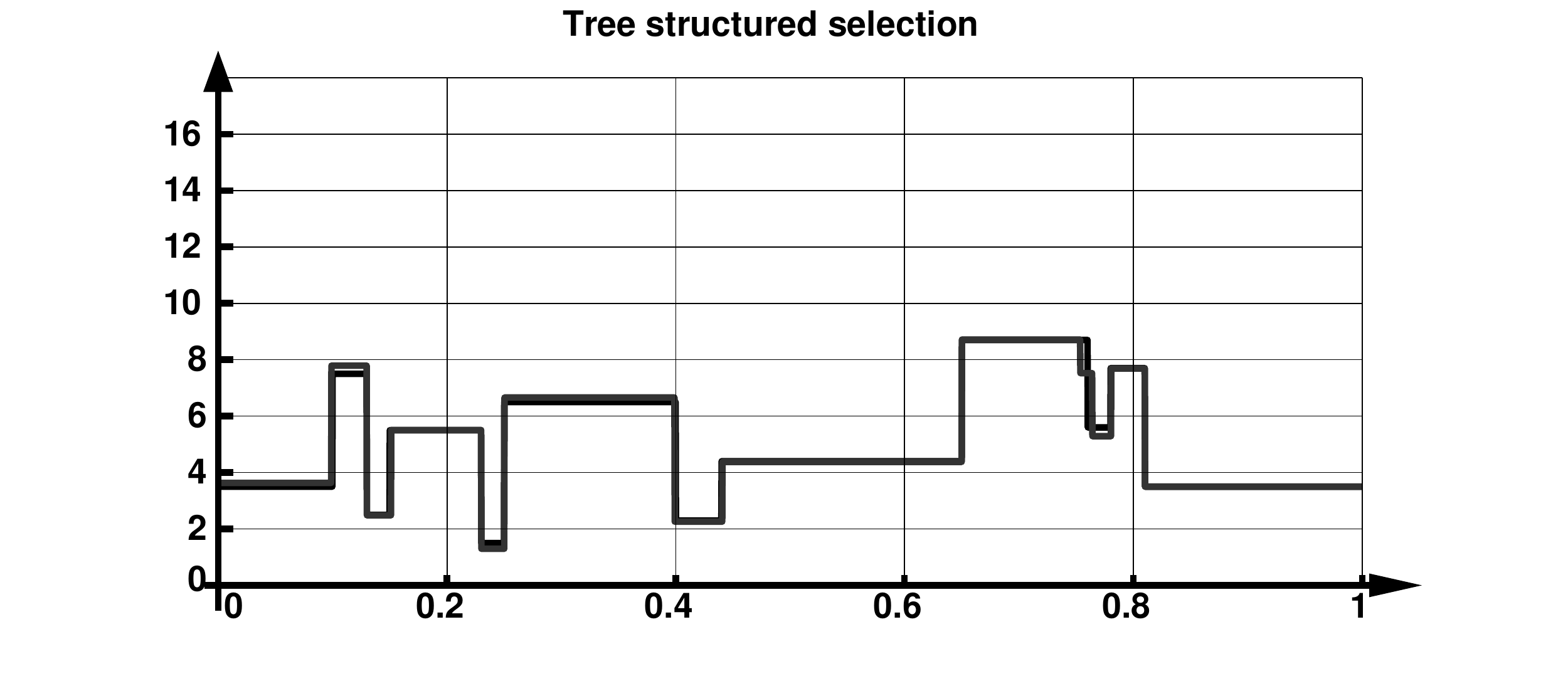}
\caption{Reconstruction of the Poisson intensity curve (black line) in Figure
\ref{fig:blockpoisson} using tree structured selection of the AUHT
coefficients of the observations (grey line in Figure \ref{fig:blockpoisson}).
}
\label{fig:blockpoissontree}
\end{center}
\end{figure}

The Figures \ref{fig:blockpoissonthr} and \ref{fig:blockpoissontree} depict two
reconstructions of the Poisson intensity curve in Figure
\ref{fig:blockpoisson}. Both reconstructions operate on an AUHT based
regression tree, in which the contrast function, used in the AUHT refinement is
given by
\[
c_{j,\ell} = {\overline{Y}_{j+1,2\ell+1} - \overline{Y}_{j+1,2\ell} \over
\sqrt{\left({1\over n_{j+1,2\ell+1}} + {1\over n_{j+1,2\ell}}\right)^q}},
\]
where $q=1$ would lead to $c_{j,\ell} = d_{j,\ell}$. Higher values of $q$
promote balanced refinements, i.e., splitting an set of points $I_{j,\ell}$
near its midpoint. It is found empirically that $q=2$ leads to better results
that $q=1$, although the issue requires a closer look in further research.

\begin{figure}[h!]
\begin{center}
\includegraphics[width=0.95\textwidth]{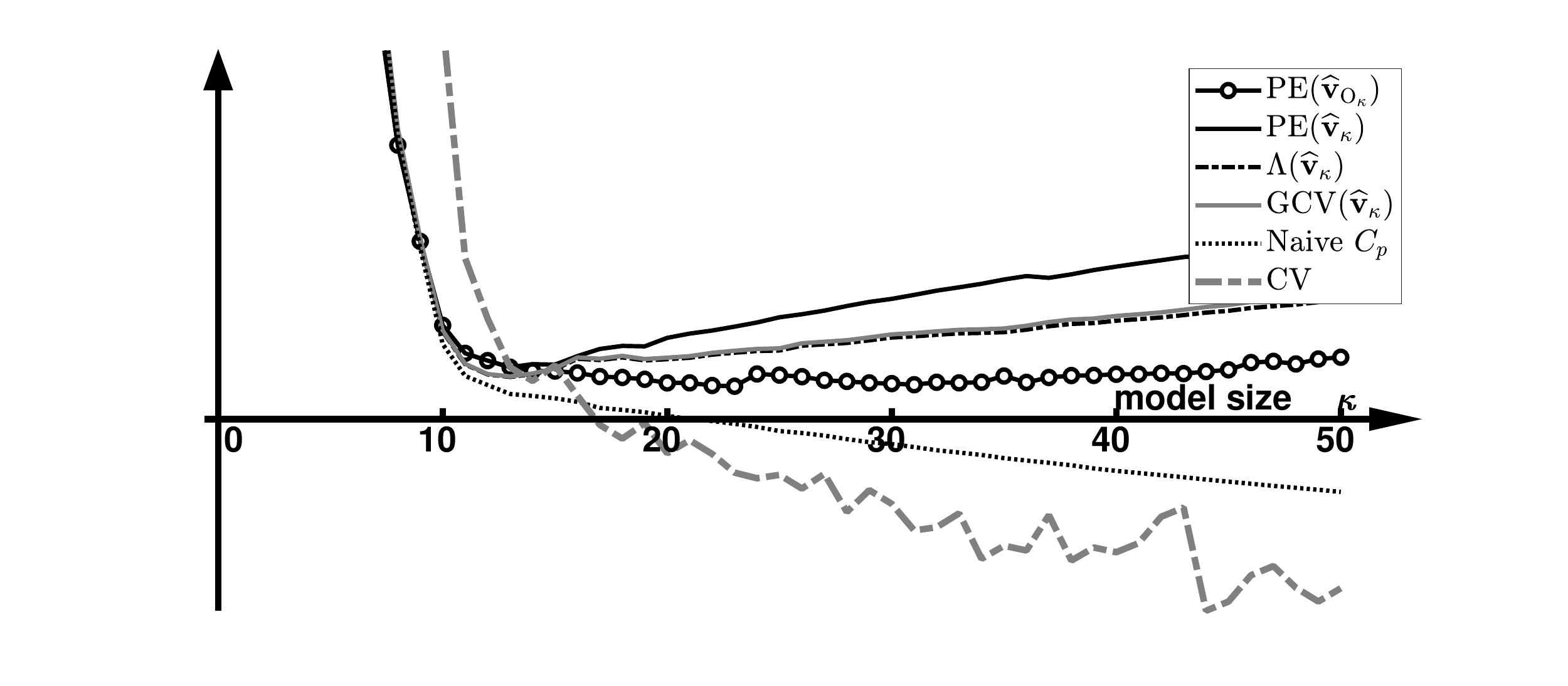}
\caption{Curve of the estimated prediction error
$\Lambda(\vec{v}_\kappa)$ as a function of the size (number of nodes) $\kappa$
of the selected subtree. For comparison, the figures also depicts the true
prediction error, $\mathrm{PE}(\widehat{\vec{v}}_\kappa)$, and the GCV
alternative for the estimated curve, using (\ref{eq:defGCV}).
Alternatives not using the mirror correction include the 
naive implementation of Mallows's $C_p$ with $\nu_\kappa = \kappa$ and
classical leave-half-out cross validation (CV). These two methods do not come
close to identifying the correct optimal $\kappa$, leading to overestimated
subtrees.
The curve of $C_p$ as a function of $\kappa$ can be seen to be the reflection
of the $\Lambda(\vec{v}_\kappa)$ curve w.r.t.~mirror, i.e., the oracular
PE-curve, $\mathrm{PE}(\widehat{\vec{v}}_{\mathrm{O}_\kappa})$.}
\label{fig:treemirror50}
\end{center}
\end{figure}

The reconstruction in Figure \ref{fig:blockpoissonthr} adopts simple,
unstructured soft-thresholding on the AUHT coefficients, where the threshold
is chosen by minimisation of the GCV expression in (\ref{eq:defGCV}).
The degrees of freedom in a soft-threshold scheme are given by 
$\nu_\kappa = \kappa$
\citep{zou07:doflasso,tibshirani12:doflasso,jansen15:gcv}.
The reconstruction in Figure \ref{fig:blockpoissontree} is obtained from the
best $\kappa$-subtree selection of the AUHT coefficients. The value of $\kappa$
is found by minimisation of $\Lambda(\vec{v}_\kappa)$ in (\ref{eq:Cptree}).
This minimisation is visualised in Figure \ref{fig:treemirror50}, which
compares several criteria for the evaluation of a best $\kappa$-subtree.
Leave-half-out Cross Validation, marked as CV in the figure and reported as an
appropriated option for applying cross validation in the context of variable
selection \citep{yang07:consistencyCV} clearly minimises at too large subtrees
(even beyond the range depicted in the figure, not to mention the fluctuations
in the curve). Naive use of Mallows's $C_p$ (with $\nu_\kappa = \kappa$) leads
to a similar behaviour. 
The plots of CV and $C_p$ in Figure \ref{fig:treemirror50} are representative
for the essential problem that affects any parameter selection (not just tree
structured approaches) based on information criteria: right after the initial,
straightforward selection of the most prominent covariates, characterised by a
steep drop of the criterion's value, the selection procedure enters a more
critical phase, in which it has to distinguish among more questionable
candidates. In this phase, any information criterion will encounter
insignificant candidates with an accidentally high score. This high score comes
from the fact that the false positive covariate carries more noise than an
arbitrary insignificant covariate. This discrepancy is described by the mirror
correction: a false positive covariate may appear to be the best candidate for
selection, whereas in reality it induces more noise than the acceptance of a
random candidate.

\begin{figure}[!]
\begin{center}
\includegraphics[width=0.95\textwidth]{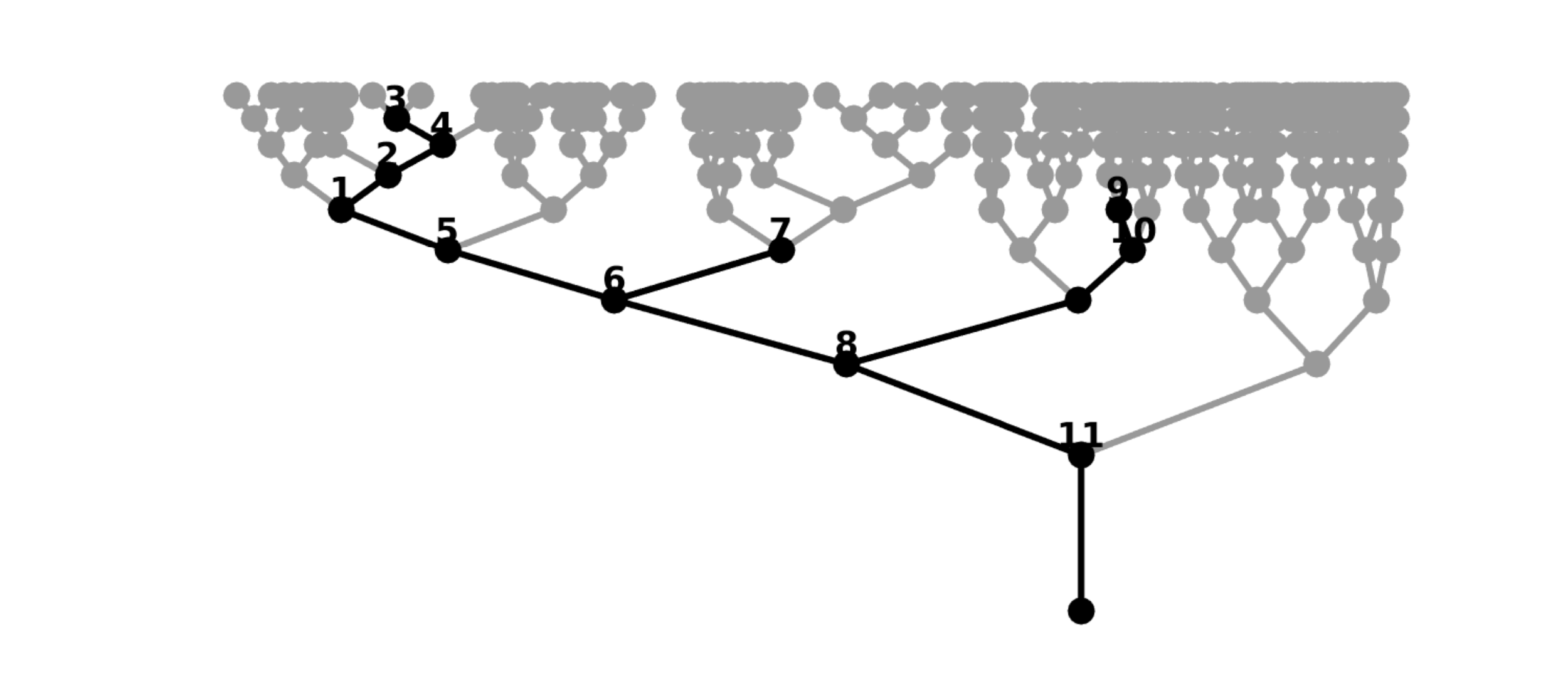}
\caption{Subtree selected by minimising $\Lambda(\widehat{v}_\kappa)$ in Figure
\ref{fig:treemirror50}, leading to the reconstruction $\widehat{\vec{\mu}}$ in
Figure \ref{fig:blockpoissontree}. The first ten levels of the full AUHT
tree are depicted in background grey.}
\label{fig:selectedtree}
\end{center}
\end{figure}

The minimisation of the $\Lambda(\widehat{v}_\kappa)$ as a function
of the subtree size $\kappa$ leads to the subtree depicted in Figure
\ref{fig:selectedtree}, which in its turn gives rise to the reconstruction in
Figure \ref{fig:blockpoissontree}.
The selected subtree contains the root of the AUHT refinement tree, which
represents to overall average value. It also contains twelve nodes
corresponding to refinements in the construction of the regression tree.
Eleven of these twelve refinements can be associated to a real change point in
the intensity curve, as can be seen from the nodes being marked with the
corresponding change point number in Figure \ref{fig:blockpoisson}. One
refinement does not correspond to a real change point, making it a false
positive. As the corresponding node fathers two real change point nodes, this
false positive is due to the construction of the AUHT, not to the tree
structured selection algorithm, nor to the finetuning of that selection
based on the minimisation of the information criterion. Also note that none of
the eleven real change points were missed.

\begin{figure}[ht!]
\begin{center}
\includegraphics[width=0.95\textwidth]{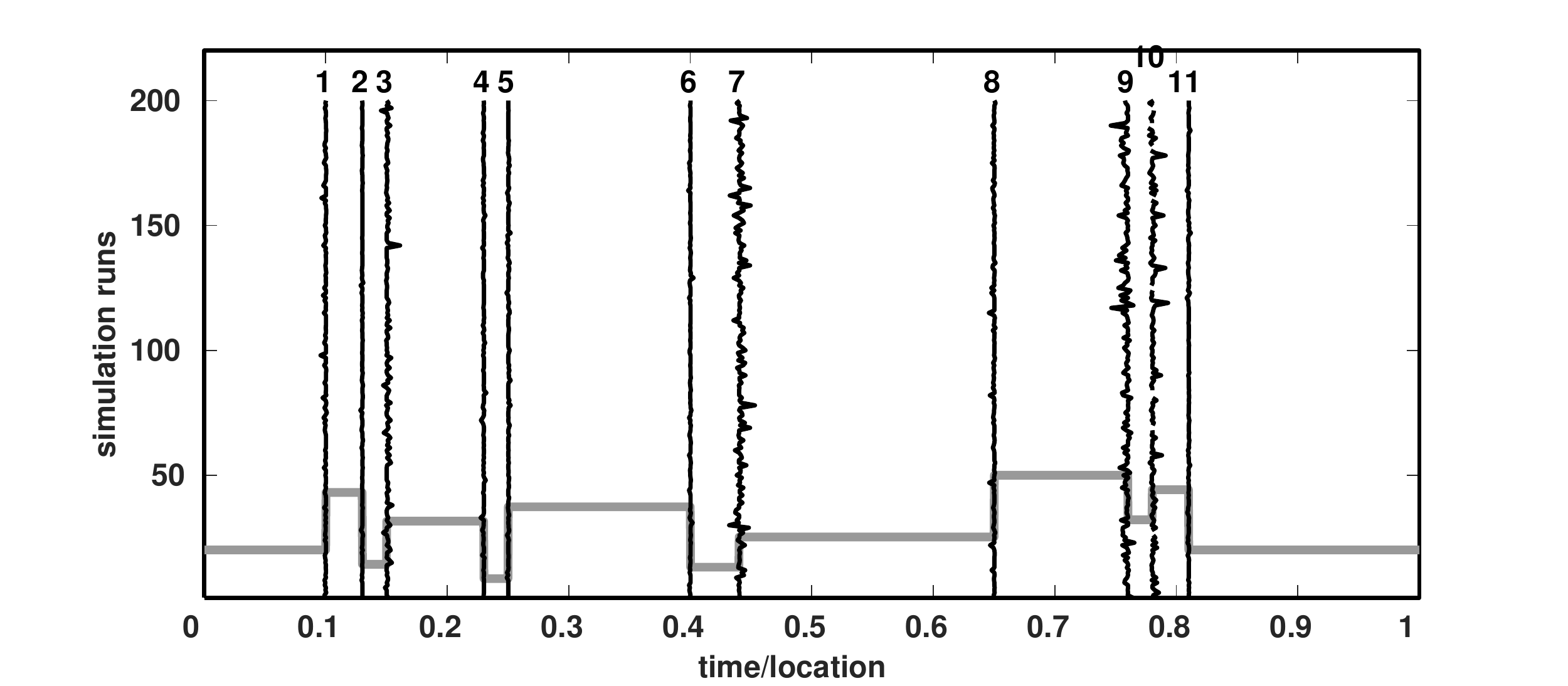}
\caption{Simulation study of 200 runs, plotting for each change point the
locations of its estimator throughout the 200 runs.}
\label{simulationsAUHTchangepoints}
\end{center}
\end{figure}

Figure \ref{simulationsAUHTchangepoints} and Table \ref{table:simulations}
summarises a simulation study for 200 realisations with the same blocks signal
intensity curve. The Figure plots the positions of the true positives across
the simulation runs, leaving a gap whenever the change point was missed in a
simulation runs (these gaps occurring mainly on the curve of change point
number 10). The plot reveals, not surprisingly, that the false negative
probability (i.e., the gap probability), as well as the variance of the
estimated location (i.e., the fluctuation of the curve) of a change point
depends on the height of the change and on the range on both sides of the
change point (i.e., the distance to the nearest change point on the left and
the right).
\begin{table}[ht!]
\begin{center}
\begin{tabular}{r|rrr}
& \multicolumn{3}{c}{Proportion of mis-} \\
& \multicolumn{3}{c}{sing change points} \\
& 0 & 1 & total \\
\hline
\multirow{6}{*}{\rotatebox[origin=c]{90}{Proportion of}}
\multirow{6}{*}{\rotatebox[origin=c]{90}{false positives}}
0 & 30.0 & 1.5 & 31.5 \\
1 & 33.5 & 3.0 & 36.5 \\
2 & 19.0 & 0   & 19.0 \\
3 &  7.0 & 0   &  7.0 \\
4 &  4.5 & 0   &  4.5 \\
5 &  1.5 & 0   &  1.5 \\
total & 95.5 & 3.4 & 100
\end{tabular}
\end{center}
\caption{Percentages of reconstructions subdivided according to number of
missing change points and number of false positives.}
\label{table:simulations}
\end{table}

\section{Sparse graphical model selection}
\label{sec:graph}

\subsection{Model and estimation}

In the second application, the information, i.e., the parameters to be
selected, estimated, and inferred, are not situated at the nodes of a tree but
rather in the edges of a graphical model. The graphical model represents the
concentration or precision matrix of a large multivariate normal random
variable,
\(
\vec{X} \sim N(\vec{\mu},\mathbf{\Sigma}).
\)
The concentration matrix is the inverse of the covariance matrix, 
$\mathbf{K} = \mathbf{\Sigma}^{-1}$, assuming the regularity of that matrix.
The concentration matrix describes the conditional dependencies
between the components of $\vec{X}$. Indeed, let $X_c$ with $c \in
\{1,2,\ldots,m\}$ be one of the components of $\vec{X}$ and denote by $c'$ the
complementary set of indices, i.e., $c'$ is $\{1,2,\ldots,m\}$ without $c$.
Furthermore, let $Y_c$ denote the observation of $X_c$ conditioned on the other 
values $\vec{X}_{c'}$, i.e., $Y_c = X_c|\vec{X}_{c'}$. Then it holds that
\begin{equation}
Y_c \sim
N(-\mathbf{K}_{c,c}^{-1}\mathbf{K}_{c,c'}(\vec{X}_{c'}-\vec{\mu}_{c'}),
\mathbf{K}_{cc}^{-1}).
\label{eq:concentration}
\end{equation}
In the subsequent discussion, we assume that $\vec{\mu}$, which is of no
interest in the question of conditional dependencies, is known to be zero.
Learning the concentration matrix can be identified as a so-called
nodewise regression problem \citep{meinshausen06:graphs,zhou11:highdimcovar}
$Y_c = \vec{X}_{c'}^T\vec{\beta}_c+\sigma_cZ_c$, where
$\sigma_c = \mathbf{K}_{cc}^{-1/2}$ while
$Z_c = \mathbf{K}_{cc}^{1/2}(Y_c-\vec{X}_{c'}^T\vec{\beta}_c)$
is a standard normally distributed random variable and
\(
\vec{\beta}_c^T = -\mathbf{K}_{c,c}^{-1}\mathbf{K}_{c,c'}.
\)
Repeated observations $\vec{X}_i$, $i=1,2,\ldots n$, of the $m$-variate 
$\vec{X}$ define for each component $c$ a $n \times (m-1)$ design matrix
$\mathbf{X}_{c'}^T$, each row corresponding to one observation in the
regression model.

With a samplesize $n$ larger than $m$, it can be hoped that the
sample covariance matrix
\[
\widehat{\mathbf{\Sigma}} = {1 \over n} \sum_{i=1}^n \vec{X}_i\vec{X}_i^T
\]
has full rank, so its inverse
\(
\widehat{\mathbf{K}} = \widehat{\mathbf{\Sigma}}^{-1}
\)
may serve as an estimator of the concentration matrix. In any case, only a
non-singular $\widehat{\mathbf{K}}$ can be the maximum likelihood estimator of
$\mathbf{K}$ in the multivariate normal model. Indeed, the log-likelihood
is given by
\[
\log\mathrm{L}(\mathbf{K})
=
\sum_{i=1}^n\left[{1 \over 2} \log\det\mathbf{K} - {m\over 2}\log(2\pi)
- {1\over 2}\vec{X}_i^T\mathbf{K}\vec{X}_i\right],
\]
which is unbounded if $\det\mathbf{K} = 0$.
As
\[
\sum_{i=1}^n\vec{X}_i^T\mathbf{K}\vec{X}_i
=
\sum_{i=1}^n\mathrm{Tr}\left(\vec{X}_i^T\mathbf{K}\vec{X}_i\right)
=
\sum_{i=1}^n\mathrm{Tr}\left(\mathbf{K}\vec{X}_i\vec{X}_i^T\right)
=
\mathrm{Tr}\left(\mathbf{K}\sum_{i=1}^n\vec{X}_i\vec{X}_i^T\right),
\]
the log-likelihood can be written as
\(
\log\mathrm{L}(\mathbf{K})
=
{n \over 2}
\left[
\log\det\mathbf{K}-\mathrm{Tr}\left(\mathbf{K}\widehat{\mathbf{\Sigma}}\right)
-m\log(2\pi)\right].
\)
A local maximum is reached in $\widehat{\mathbf{K}}_\mathrm{ML}$ if
\(
\nabla\log\mathrm{L}\left(\widehat{\mathbf{K}}_\mathrm{ML}\right)
\)
equals the zero matrix.
As $\nabla\log\det\mathbf{K} = \mathbf{K}^{-T}$ and
\(
\nabla \mathrm{Tr}\left(\mathbf{K}\widehat{\mathbf{\Sigma}}\right)
=
\widehat{\mathbf{\Sigma}}^T,
\)
this can be verified to develop as
\(
\widehat{\mathbf{K}}_\mathrm{ML}\widehat{\mathbf{\Sigma}} = \mathbf{I}.
\)
The solution provided by nodewise regression \citep{meinshausen06:graphs}
satisfies this equation if $\widehat{\mathbf{\Sigma}}$ is non-singular,
but not otherwise. Nodewise regression does not impose symmetric concentration
matrices explicitly, although symmetry of $\widehat{\mathbf{K}}_\mathrm{ML}$
follows automatically whenever $\widehat{\mathbf{K}}_\mathrm{ML}$ is indeed the
maximum likelihood estimator.
In practice, the computation of $\widehat{\mathbf{K}}_\mathrm{ML}$ is often
unstable, even when $n$ is larger than $m$. The graphical lasso 
\citep{banerjee08:multivariate,friedman08:graphicallasso,mazumder12:graphicallasso,sojoudi16:glassothr}
obtains a regularised estimator $\widehat{\mathbf{K}}_\lambda$ by the
maximisation
\[
\widehat{\mathbf{K}}_\lambda = \arg\max_{\mathbf{K}}
\left[
\log\det\mathbf{K}-\mathrm{Tr}\left(\mathbf{K}\widehat{\mathbf{\Sigma}}\right)
- \lambda \norm{\mathbf{K}}_1\right],
\]
where
$\norm{\mathbf{K}}_1 = \mathrm{Tr}\left(\mathbf{K}\sign(\mathbf{K})^T\right)$
stands in this case for the sum of the absolute values of all elements in
$\mathbf{K}$ (hence not for the classical induced $\ell_1$ matrix norm).
Solvers for this constrained optimisation problem have been proposed based on
an iterative sequence of lasso solvers, where each iteration step works on one
column of $\widehat{\mathbf{\Sigma}}_\lambda$, keeping the other columns
constant in that iteration step. The iterative solving increases the
computational complexity, compared to nodewise regression
\citep{meinshausen06:graphs}, which is an issue if we want to equip the solver
with a finetuning of the regularisation. In applications involving big data,
the nodewise regression procedure is easy to implement on parallel computers.
Moreover, while in a simple lasso problem the link between the regularisation
parameter $\lambda$ and the size $\kappa$ of the active set is easy to
establish, this problem is nontrivial in the framework of graphical models for
sparse concentration matrices.
For these reasons, we adopt the direct solver of \citet{meinshausen06:graphs}
as selection method. Once the set of nonzeros in the concentration matrix has
been selected, estimation within this set takes place according to a 
constrained maximum likelihood principle, as outlined in the following section.

\subsection{Estimation of the nonzero elements in the concentration matrix}

Let $\mathrm{S}_\mathrm{NW} \subset \{1,2,\ldots,m\}\times\{1,2,\ldots,m\}$
be the selection by the lasso in the nodewise regression framework.
A pair $(i,j) \in \mathrm{S}_{\mathrm{NW}}$ means that the corresponding 
entry in the estimated concentration matrix is nonzero.
The selection proceeds row by row by application of lasso to the vectors
\(
\vec{\beta}_c^T = -\mathbf{K}_{c,c}^{-1}\mathbf{K}_{c,c'}
\)
with $c \in \{1,2,\ldots,m\}$ in the linear models (\ref{eq:concentration}).
The selection is finetuned by optimisation of the criterion
\[
\widehat{\Lambda}_c(\widehat{\vec{\beta}}_{c\kappa})
=
{1 \over n} \mathrm{SS_E}(\widehat{\vec{\beta}}_{c\kappa}) 
               + {2\kappa \over n} \sigma^2 + 2\widehat{m}_\kappa - \sigma^2,
\]
i.e., by filling in (\ref{eq:defmirror}) into (\ref{eq:defCpnonstud}), and
estimating the correction as in (\ref{eq:EnormPSgivenS}).
As before, a pilot estimator can be used to deal with the nuisance parameter
$\sigma^2$. The nodewise lasso offers no automatic symmetry in the selection.
The subsequent discussion assumes the selection $\mathrm{S}_\mathrm{NW}$ to be
symmetrised by keeping the pair $(i,j)$ in $\mathrm{S}_\mathrm{NW}$ if and
only if $(j,i)$ was also selected.

Once $\mathrm{S}_\mathrm{NW}$ has been identified, the shrinkage estimator of
the lasso is replaced by the constrained maximum likelihood estimator
$\widehat{\mathbf{K}}_{\mathrm{ML},\mathrm{S}}$, maximising
$\log\mathrm{L}(\mathbf{K})$ under the condition that
$\widehat{K}_{\mathrm{ML},\mathrm{S};ij} = 0$ unless
$(i,j) \in \mathrm{S}_\mathrm{NW}$.
Introducing a matrix of Lagrange multipliers
$\widehat{\mathbf{\Lambda}}_{\mathrm{S}}$ with
$\widehat{\Lambda}_{\mathrm{S};ij} = 0$ whenever
$(i,j) \in \mathrm{S}_\mathrm{NW}$ (for the corresponding element of
$\widehat{\mathbf{K}}_{\mathrm{ML},\mathrm{S}}$
is unconstrained), the constrained maximum likelihood problem is given by
\[
\widehat{\mathbf{K}}_{\mathrm{ML},\mathrm{S}} = \arg\max_{\mathbf{K}}
\left[
\log\det\mathbf{K}-\mathrm{Tr}\left(\mathbf{K}\widehat{\mathbf{\Sigma}}\right)
- \mathrm{Tr}\left(\mathbf{K}\widehat{\mathbf{\Lambda}}_{\mathrm{S}}\right)
\right].
\]
Taking the derivatives w.r.t.~the elements in $\mathbf{K}$ leads to
\(
\mathbf{K}^{-T}-\widehat{\mathbf{\Sigma}}^T-\widehat{\mathbf{\Lambda}}_{\mathrm{S}}^T
= \vec{0}.
\)
The condition that $\widehat{\Lambda}_{\mathrm{S};ij} = 0$ for $(i,j) \in
\mathrm{S}_\mathrm{NW}$ then means that the gradient
\(
\mathbf{\Lambda}_{\mathrm{S}}^T = \nabla\log\mathrm{L}\left(\mathbf{K}\right) =
\mathbf{K}^{-T}-\widehat{\mathbf{\Sigma}}^T
\)
must have zero entries in all $(i,j) \in \mathrm{S}_\mathrm{NW}$.

Unless $\mathrm{S}_\mathrm{NW}$ contains all possible pairs $(i,j)$,
nodewise regression cannot possibly find this constrained maximum likelihood
solution. Indeed, Let $(i,j) \not\in \mathrm{S}_\mathrm{NW}$, then
$\widehat{K}_{ij} = \widehat{K}_{ji}$ is imposed to be zero. This affects all
other $\widehat{K}_{ik}$ and $\widehat{K}_{jk}$ at rows $i$ and $j$ of
$\widehat{\mathbf{K}}$.
Nodewise regression at row $c=k$, unaware of the zero $\widehat{K}_{ij}$, will
find values $\widehat{K}_{ki}$ and $\widehat{K}_{kj}$ as if there is no
constraint on $\widehat{K}_{ij}$. The resulting estimated
concentration matrix cannot be symmetric (even though the selection
$\mathrm{S}_\mathrm{NW}$ is symmetric), nor can it maximise the likelihood.
Let $\widehat{\mathbf{K}}_{\mathrm{NW},\mathrm{S}}$ be the outcome of
constrained nodewise regression, then a symmetrised version of it, for instance
\(
(\widehat{\mathbf{K}}_{\mathrm{NW},\mathrm{S}}+\widehat{\mathbf{K}}_{\mathrm{NW},\mathrm{S}}^T)/2,
\)
can be used as initial value in an iterative search for
\(
\widehat{\mathbf{K}}_{\mathrm{ML},\mathrm{S}}.
\)
An iterative search can be implemented by projection of the gradient
$\mathbf{\Lambda}_{\mathrm{S}}^T$ onto the space of admissible descends,
i.e., replacing the partial derivative w.r.t.~$K_{ij}$ by zero if
$(i,j) \not\in \mathrm{S}_\mathrm{NW}$. The iteration stops as soon as the
other partial derivatives, those w.r.t.~elements in $\mathrm{S}_\mathrm{NW}$
are zero. 

Let $\mathbf{\Lambda}_{\mathrm{S},0}^T$ denote the search direction, then the
iteration step updates the current solution $\mathbf{K}$ to
$\mathbf{K}+\omega\mathbf{\Lambda}_{\mathrm{S},0}^T$, where $\omega$ maximises
the function
\(
g(\omega) =
\log\mathrm{L}\left(\mathbf{K}+\omega\mathbf{\Lambda}_{\mathrm{S},0}^T\right).
\)
Taking the derivative yields 
\[
g'(\omega) =
\mathrm{Tr}\left[\left(\mathbf{I}+\omega\mathbf{K}^{-1}
\mathbf{\Lambda}_{\mathrm{S},0}^T\right)
\mathbf{K}^{-1}\mathbf{\Lambda}_{\mathrm{S},0}^T\right]
-
\mathrm{Tr}\left[\widehat{\mathbf{\Sigma}}\mathbf{\Lambda}_{\mathrm{S},0}^T\right].
\]
With $\vec{\gamma}$ denoting the vector of eigenvalues of $\mathbf{K}^{-1}
\mathbf{\Lambda}_{\mathrm{S},0}^T$, this is
\[
g'(\omega) = \sum_{i=1}^m {\gamma_i \over 1+\omega\gamma_i} -
\mathrm{Tr}\left[\widehat{\mathbf{\Sigma}}\mathbf{\Lambda}_{\mathrm{S},0}^T\right],
\]
whose zero can be found numerically.

\subsection{A short simulation study}

\begin{figure}
\begin{center}
\begin{tabular}{cc}
\includegraphics[width=0.49\textwidth]{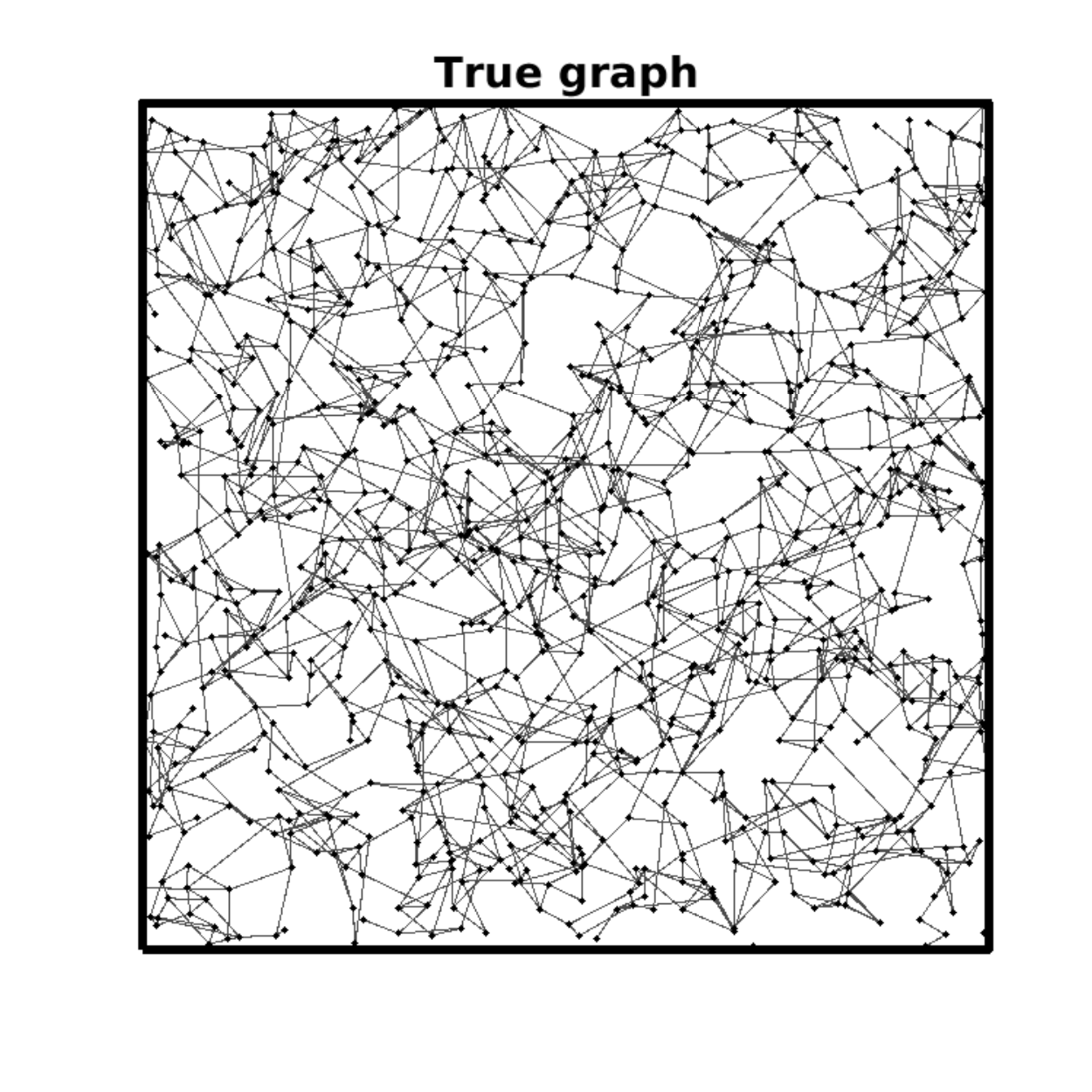} &
\includegraphics[width=0.49\textwidth]{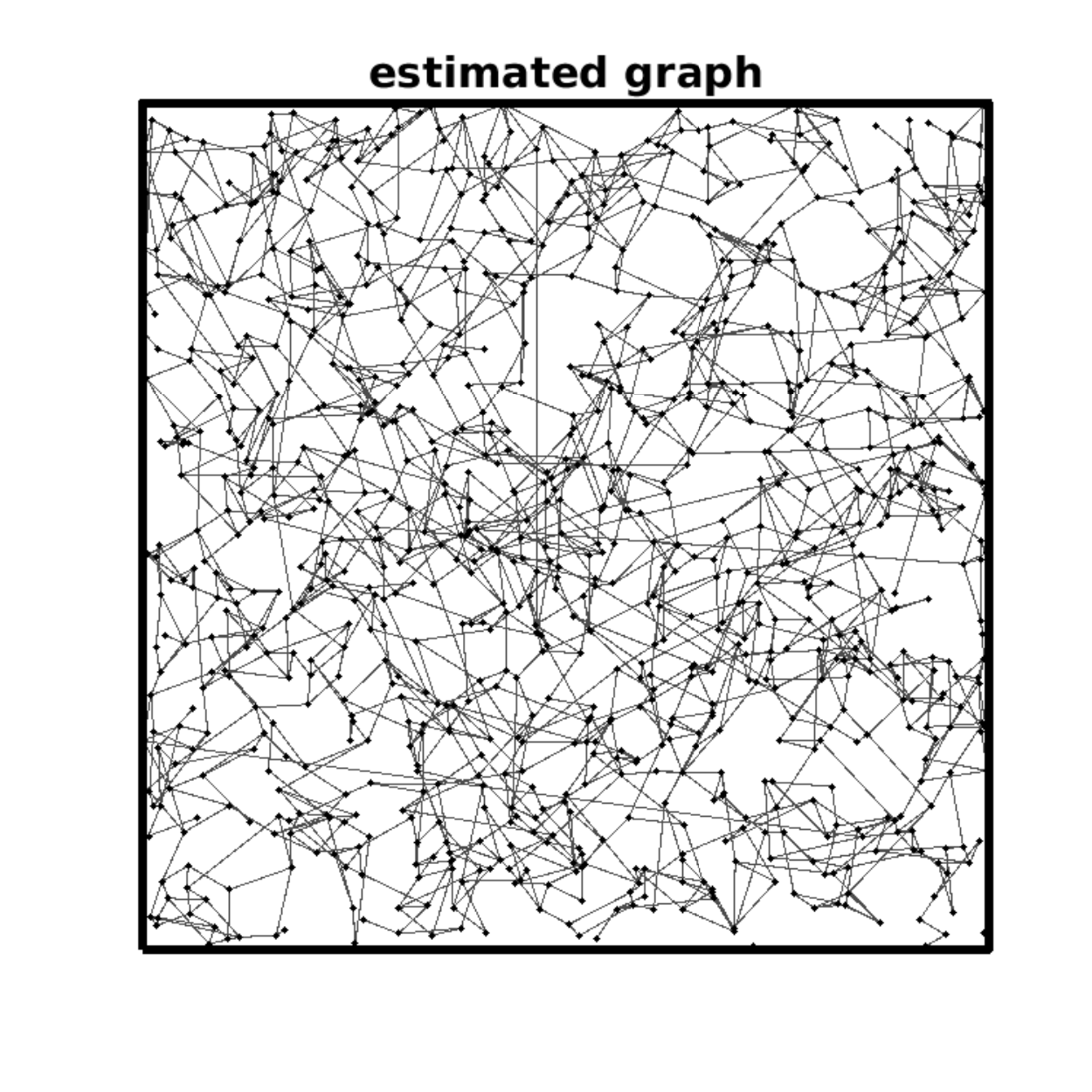} \\
(a) & (b) \\
\multicolumn{2}{c}{
\includegraphics[width=0.49\textwidth]{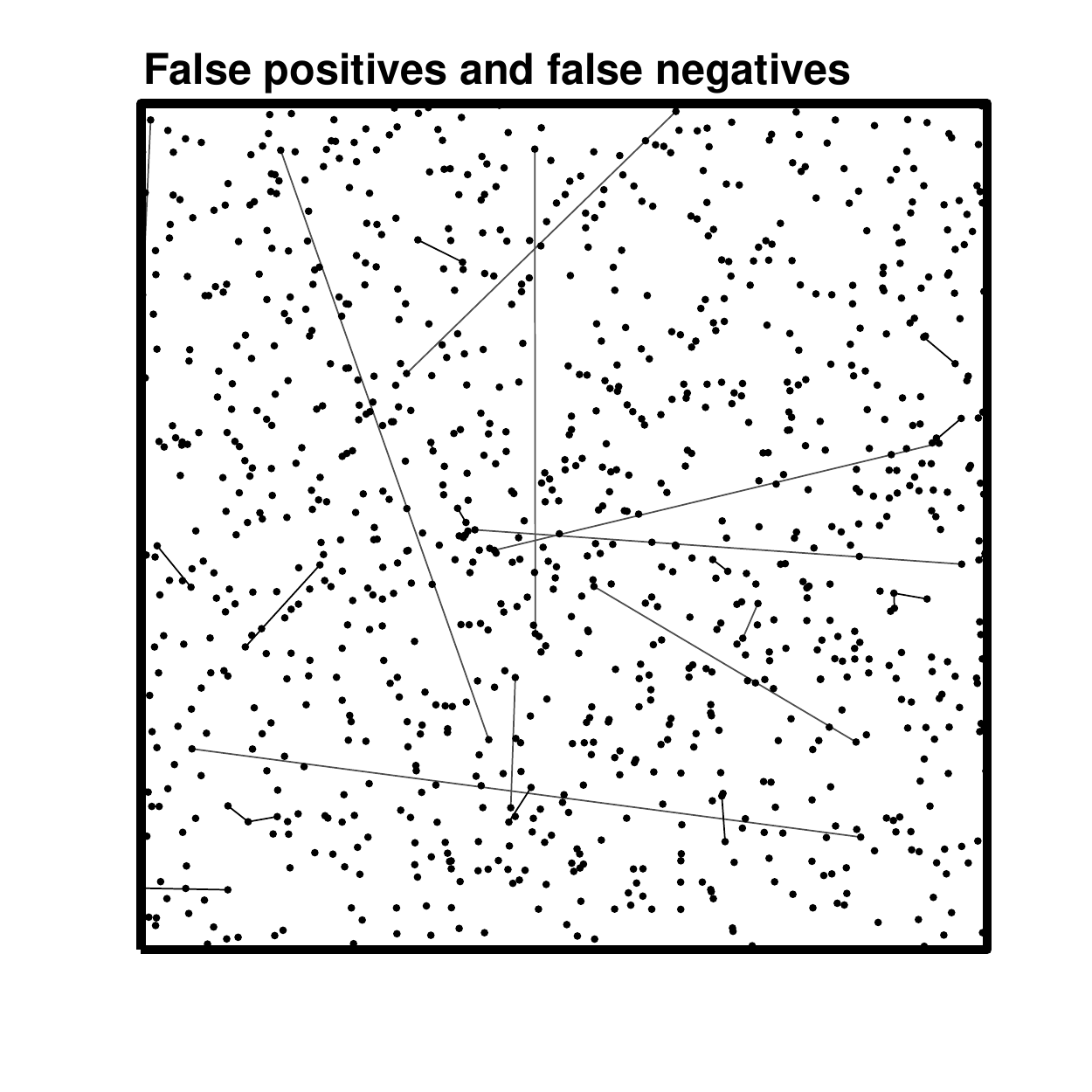}
}
\\
\multicolumn{2}{c}{(c)}
\end{tabular}
\caption{Simulation of a nodewise regression on a sample of $n=600$
observations of $m$-variate normal random vector, with $m=1000$, as in
\citet{meinshausen06:graphs}. (a) Graph representing the sparse concentration
matrix $\mathbf{K}$. (b) Estimation of the graph based on the proposed refined
Mallows's $C_p$ criterion. (c) False positives and false negatives (type I and
type II errors). The false positives are the long edges, connecting
components $i$ and $j$ far from each other in the scatter plot. (The estimation
method is not aware of the distances in the scatter plot.)}
\label{fig:simulationnodewise}
\end{center}
\end{figure}

Before proceeding to a real data analysis, a short simulation study reveals
some understanding in the working of the proposed sparse selection method.
Figure \ref{fig:simulationnodewise} displays the setting of the
simulation study of \citet[p.1448]{meinshausen06:graphs}, along with the
outcome of the proposed refined Mallows's $C_p$ criterion in nodewise
regression.
\begin{figure}[t!]
\begin{center}
\begin{tabular}{cc}
\includegraphics[width=0.49\textwidth]{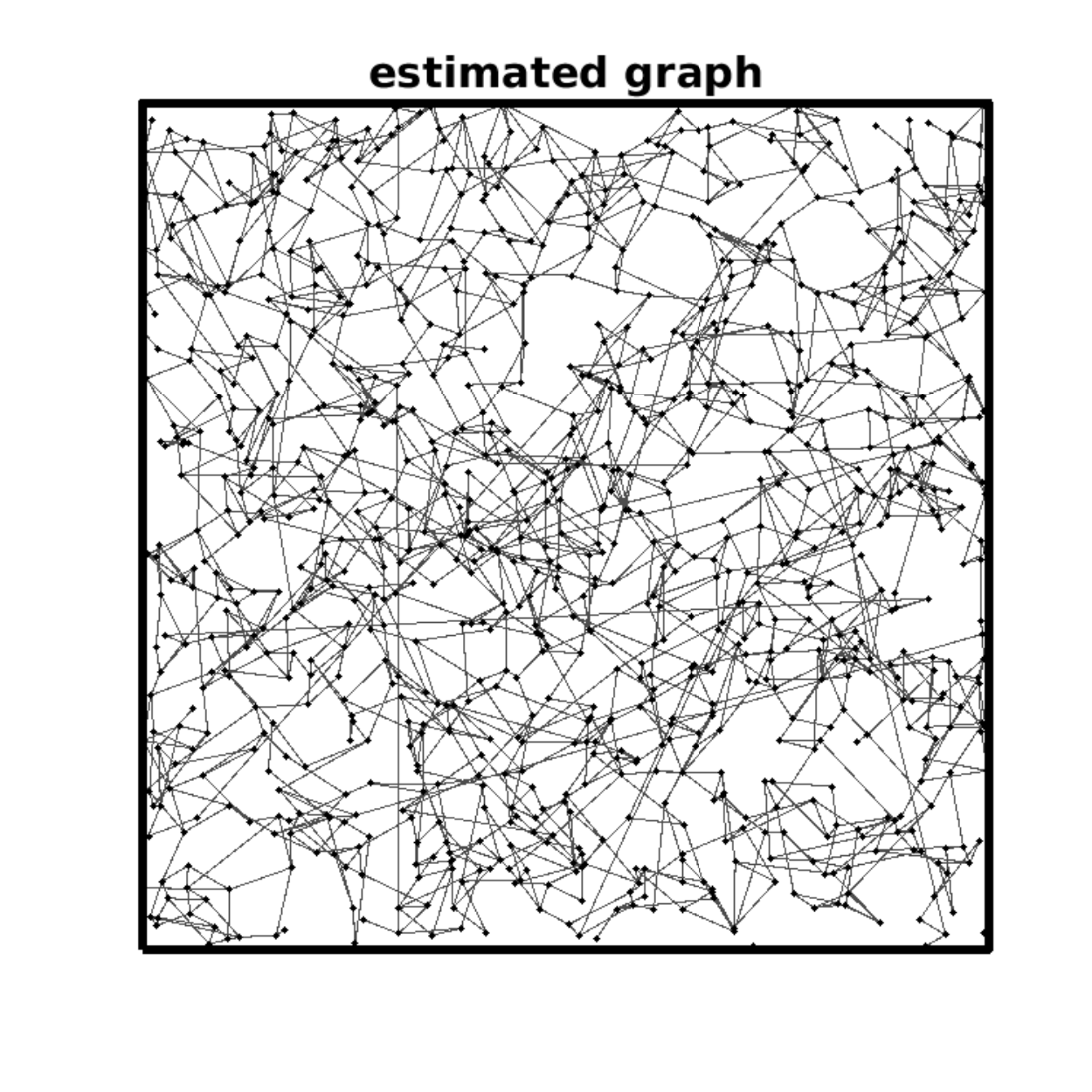} &
\includegraphics[width=0.49\textwidth]{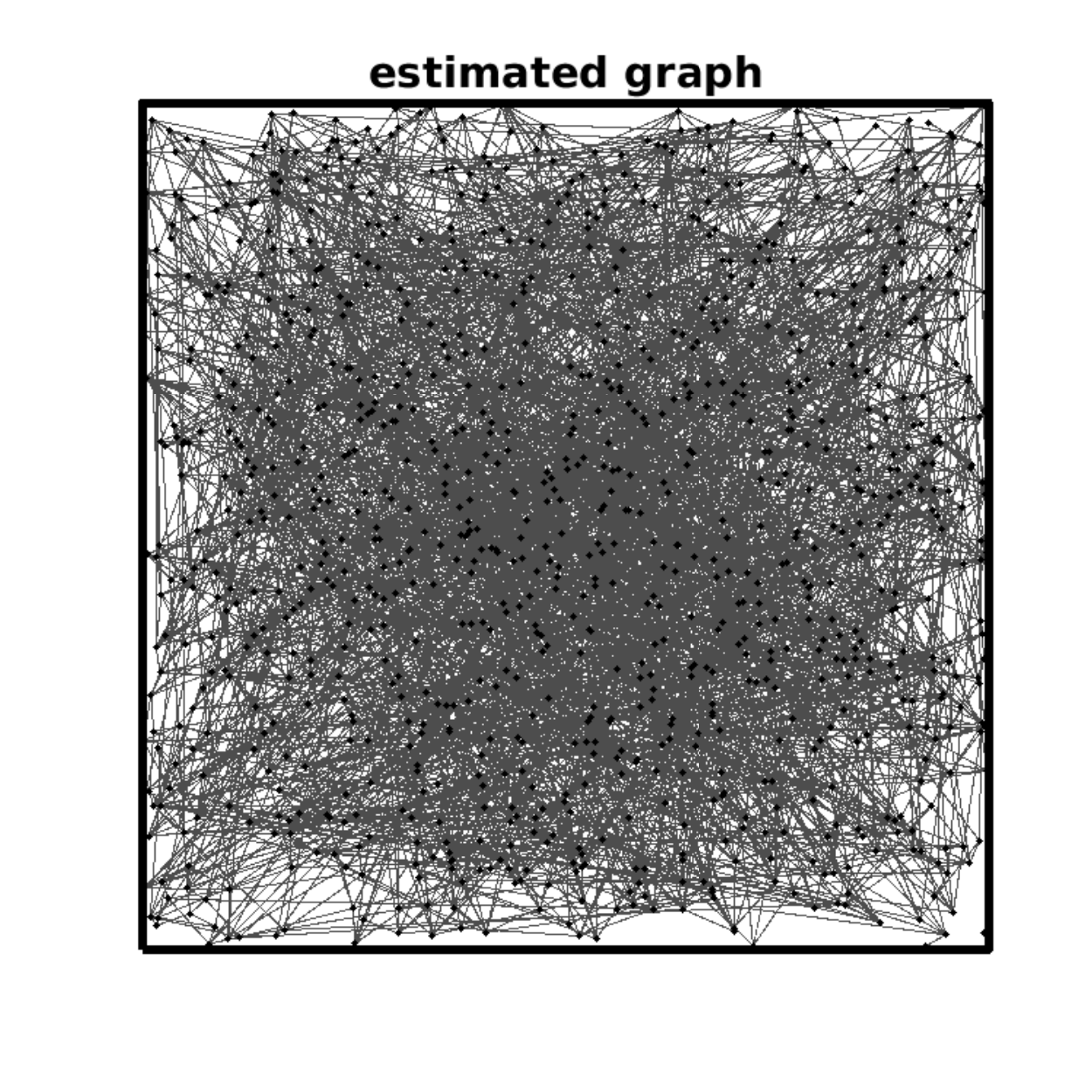} \\
(a) & (b)
\end{tabular}
\caption{(a) Lasso shrinkage estimation using the oracular information that
the number of edges in each node is bounded by four.
(b) Nodewise regression using Lasso shrinkage implemented
by LARS with Mallows's $C_p$ as stopping criterion.}
\label{fig:simulationnodewise2}
\end{center}
\end{figure}
The results in the figure were obtained from a sample of $n=600$ independent
observations of a $m$-variate normal random vector $\vec{X}$, with $m=1000$.
For the sake of the graphical representation, the components of $\vec{X}$ are
associated with random nodes in a 2D scatter plot, depicted in the Figure.
The positions of the nodes have no physical meaning, except for their role 
in the generation of the concentration matrix.
The concentration matrix $\mathbf{K}$ is sparse in the sense that at most
$n_e = 4$ nonzero off-diagonals appear on each row and each column.
Nonzero $K_{ij}$ are selected at random, but with probabilities roughly
inversely proportional to the distances between the nodes corresponding to the
$i$th and $j$th components of $\vec{X}$. The graph representing $\mathbf{K}$
will therefore show short edges between adjacent nodes, as can be seen in
Figure \ref{fig:simulationnodewise}(a).
Initially, the values of the nonzero off-diagonals in $\mathbf{K}$ are set to
$(1/n_e-0.005) = 0.245$ to guarantee positive definiteness of $\mathbf{K}$.
After inversion of the initial matrix $\mathbf{K}$, the resulting covariance
matrix is standardised by left- and right multiplication with a diagonal matrix
so that the diagonal elements of the covariance matrix (the variances, that is)
are all one. The concentration matrix is rescaled accordingly.

As illustrated in Figures \ref{fig:simulationnodewise}(b) and
\ref{fig:simulationnodewise}(c),
the proposed method cannot eliminate all false positives (type I errors), nor
does it prevent the occurrence of false negatives (type II errors).
Instead, it succeeds in finding a delicate balance between the two objectives.
Theoretic results quantifying these findings are interesting topics of further
research.
At first sight, the result in \citet{meinshausen06:graphs}, depicted in Figure
\ref{fig:simulationnodewise2}(a), is superior, were it not for the upperbound
of four on the number of edges in each vertex, used throughout the nodewise
regression. In some applications, knowledge of 
such an upperbound may be available, whereas in other applications, this
information should be considered as oracular. In absence of this information,
LARS equipped with the classical $C_p$ based stopping criterion (having
$\kappa$ instead of $\nu_\kappa$ as penalty in (\ref{eq:defCpnonstud})
would lead to a massive overestimation of the true model, as depicted in
Figure \ref{fig:simulationnodewise2}(b).
Alternatively, a selection with focus on the false discovery rate may be too
conservative, leading to an uncontrolled number of false negatives. 

\subsection{A real data example}

The symmetrised nodewise regression approach with the proposed refined
Mallows's $C_p$ is now applied to the gene expression measurements reported in
\citet{spira07:airway} and analysed in \citet{danaher14:glasso} to illustrate
the graphical lasso across multiple populations. The data are available from
the Gene Expression Omnibus \citep{barrett05:NCBI},
\url{https://www.ncbi.nlm.nih.gov/geo/}, accession code GDS2771.

The observations come from two populations: $n_1 = 90$ individuals belong to
the control group, while $n_2 = 97$ patients have been diagnosed with lung
cancer. The objective is to investigate whether the diagnosis explains
differences in covariance structure between the gene expressions.
Just like in \citet{danaher14:glasso}, the genes in the upper 20\% quantile
of the expression variances are taken out from further analysis, because
covariances among these genes are supposed to be dominated by noise.
The expression measurements of the remaining $m=17827$ genes (out of 22283
originally) are studentised within each population.

\begin{figure}[!]
\begin{center}
\includegraphics[width=0.5\textwidth]{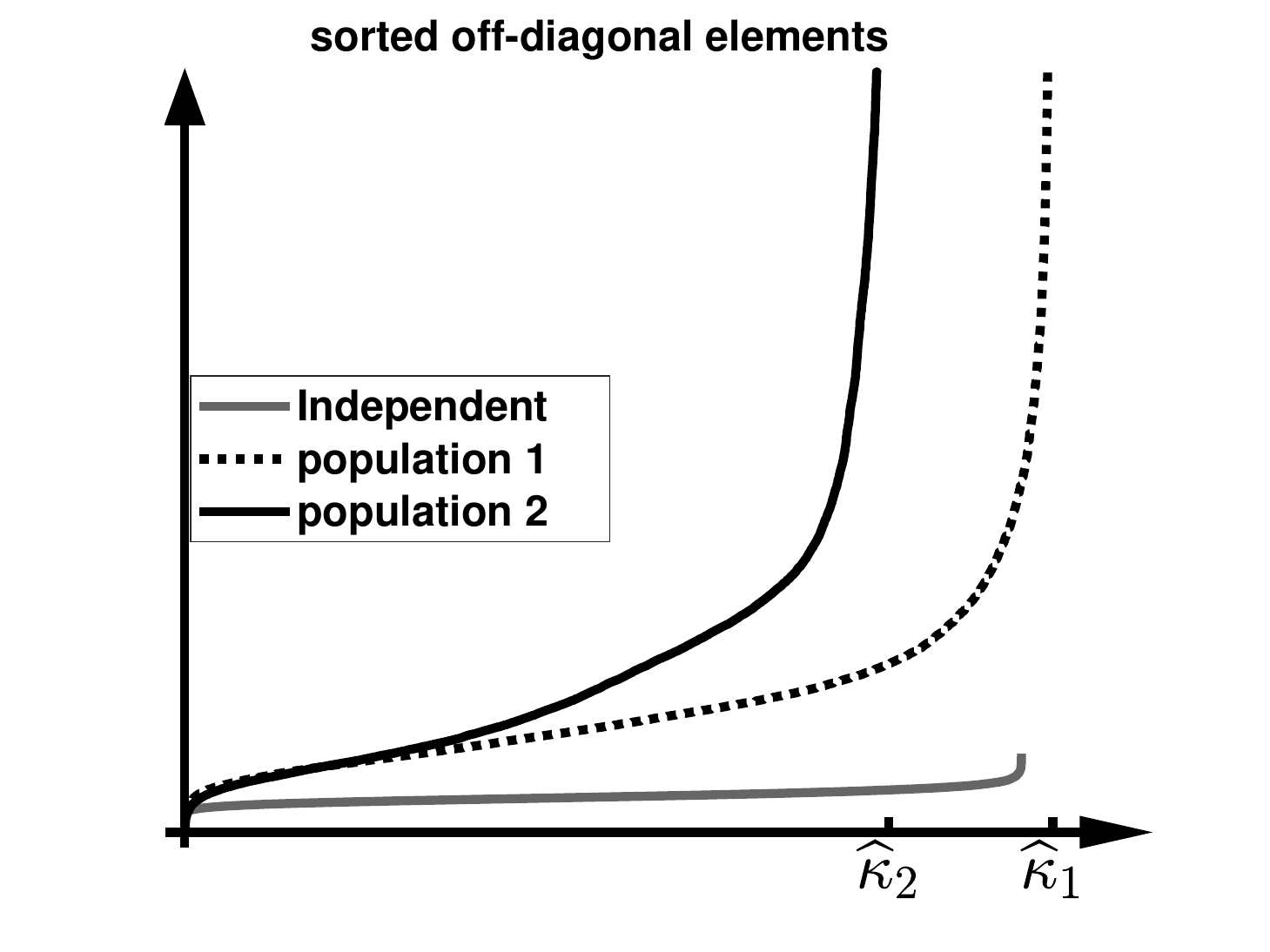}
\caption{Sorted magnitudes of the selected off-diagonal entries of the
concentration matrix. Comparison with (falsely) selected off-diagonal entries
in a simulated vector of independent random variables.}
\label{fig:sortedselected}
\end{center}
\end{figure}

Let $\vec{X}_1 \sim N(\vec{\mu}_1,\mathbf{\Sigma}_1)$ denote the gene
expressions in the control group and
$\vec{X}_2 \sim N(\vec{\mu}_2,\mathbf{\Sigma}_2)$ those of the patients,
then the symmetrised nodewise regression with the proposed refined Mallows's
$C_p$ criterion (with the same settings as in the simulation study) selects 
$\widehat{\kappa}^*_1 = 16198$ and $\widehat{\kappa}^*_2 = 13134$ nonzero
concentration values $\widehat{K}_{ij} = \widehat{K}_{ji}$, with
$1\leq i<j\leq m$. In a full model of $m(m-1)/2$ concentration parameters,
these selections account for and $0.0102\%$ and $0.0083\%$ of nonzeros
respectively. Motivated by application specific, practical considerations, the
selections in \citet[Sections 6 and 8]{danaher14:glasso} are even sparser.
It may be interesting, however, to allow a wider (yet still very sparse)
selection in a first stage, as illustrated by Figure \ref{fig:sortedselected}.
The Figure depicts the ordered magnitudes of the selected 
off-diagonal elements of the concentration matrix (the elements that are
represented by an edge in the graphical model). The values are compared with
those obtained by applying the same estimation to $n$ Monte-Carlo 
observations from a vector of independent normal random variables.
The simulated variables have the same variances as the observed ones, in the
sense that the variances in the simulation are taken from the diagonal of the
sample covariance matrix. With diagonal covariance and concentration matrices,
the graphical model of the simulated data is know to have no edges. As the
refined Mallows's $C_p$ criterion of this paper reduces false positive
selections, it can be expected that the size of the selected set is much
smaller in the simulated data than in the observed control and patient data.
Therefore, for the sake of comparison, the selection on the simulated data 
uses the classical definition of Mallows's $C_p$, by taking $\nu_\kappa =
\kappa$ in (\ref{eq:defCpnonstud}), leading to a vast selection of all false
positives. Comparing the magnitudes of these false positives with the selected
values in the observed data reveals that they are much smaller and flatter when
sorted. This suggests that the selected values in the two populations do have
some significance. 

\section{Conclusion}

Parameter selection in high dimensional data is often monitored by practical,
application driven considerations or by methods explicitly controlling the
false positives to at least some degree. Information criteria, such as
Mallows's $C_p$, but also AIC and others, are often found to be too tolerant of
false positives. This paper has explored the use of more a refined Mallows's
$C_p$ criterion in high dimensional graphical and tree models.
The classical definition of Mallows's $C_p$, designed for assessment of
a \emph{fixed} model, works well for finetuning a lasso shrinkage selection.
As lasso shrinkage is tolerant of the presence of false positives, this
finetuning leads to largely overestimated models.
In contrast to this, the refined criterion developed for graphs and trees in
this paper, focuses on \emph{finetuning} selection for estimation
\emph{without shrinkage}.
This way, the refined criterion, carefully balancing false positive and false
negative selections, proves to be interesting in applications where the
avoidance of both false positives and false negatives are important objectives.

\section{Software}

Software for the reproduction of the figures is available in the Matlab package
\emph{ThreshLab}, available on
\url{https://maarten.jansen.web.ulb.be/software/threshlab.html}.
After installation, type \texttt{help makefigsdoftreesandgraphs} to get
started.